\documentclass{article}

\usepackage{microtype}
\usepackage{graphicx}
\usepackage{subfigure}
\usepackage{booktabs} 

\usepackage{hyperref}

\usepackage[accepted]{icml2023}

\usepackage{amsmath}
\usepackage{amssymb}
\usepackage{mathtools}
 \usepackage{multirow}
\usepackage{amsthm}

\usepackage[capitalize,noabbrev]{cleveref}

\theoremstyle{plain}

\theoremstyle{definition}

\theoremstyle{remark}

\usepackage[textsize=tiny]{todonotes}

\icmltitlerunning{TECP: Token-Entropy Conformal Prediction for LLMs}

\begin{document}

\twocolumn[
\icmltitle{TECP: Token-Entropy Conformal Prediction for LLMs}

\begin{icmlauthorlist}
\icmlauthor{Beining Xu}{su}
\icmlauthor{Yongming Lu}{suu}
\end{icmlauthorlist}

\icmlaffiliation{su}{Department of School of Engineering, Shenzhen MSU-BIT University, Shenzhen, China, 518000}

\icmlaffiliation{suu}{MSU-BIT-SMBU Joint Research  Center of Applied Mathematics, Shenzhen MSU-BIT University, Shenzhen, China, 518000}
\icmlcorrespondingauthor{Yongming Lu}{luym@smubu.edu.cn}

\vskip 0.3in
]
\printAffiliationsAndNotice{\icmlEqualContribution}
\begin{abstract}
Uncertainty quantification (UQ) for open-ended language generation remains a critical yet underexplored challenge, especially under black-box constraints where internal model signals are accessible. In this paper, we introduce Token- Entropy Conformal Prediction (TECP). This novel framework leverages token-level entropy as a logit-free, reference-free uncertainty measure and integrates it into a split conformal prediction (CP) pipeline to construct prediction sets with formal coverage guarantees. Unlike existing approaches that rely on semantic consistency heuristics or white-box features, TECP directly estimates episodic uncertainty from the token entropy structure of sampled generations and calibrates uncertainty thresholds via CP quantiles to ensure provable error control. Empirical evaluations across six large language models and two benchmarks (CoQA and  TriviaQA) demonstrate that TECP consistently achieves reliable coverage and compact prediction sets, outperforming prior self UQ methods. Our method provides a principled and efficient solution for trustworthy generation in black-box LLM settings.
\end{abstract}

\section{Introduction}

Large Language Models (LLMs) are increasingly serving as the core technological substrate across diverse tasks and exhibit outstanding cross-domain capabilities~\cite{veeramachaneni2025large,lu2025mv,rong2025backdoor,chen2025does}. With the continued accumulation and effective utilization of high-quality data, their performance has consistently improved, revealing substantial application potential in healthcare, code generation, scientific research, and psychological counseling~\cite{yang2024pediatricsgpt,chen2024miss}. Represented by systems such as ChatGPT, pretrained on large-scale corpora and finely aligned with human preferences, these models not only adapt flexibly to heterogeneous demands but also significantly enhance the efficiency and reliability with which AI handles both everyday workflows and complex professional tasks~\cite{chen2024can,zhang2023spot}, laying a firm foundation for the deeper integration of intelligent technologies and sustained innovation.

Despite their strong performance across many tasks, LLMs still exhibit pervasive reliability issues~\cite{jiang2025comt,huang2025survey,lavrinovics2025knowledge,zhou2025hademif}, including hallucinations and factual errors, and may produce responses that appear well-formed yet are fabricated or detached from reality. It reduces their adaptability for deployment in high-stakes scenarios. Against this backdrop, uncertainty quantification (UQ) has become a key approach for assessing the credibility of model outputs, enabling users to determine when a model’s answers can be trusted~\cite{mora2024uncertainty,farquhar2024detecting,duan2024shifting,qiu2024semantic,wang-etal-2024-conu,wang2025word}. It is especially critical in high-risk professional domains such as medicine and psychology, where decision-making often relies heavily on LLM-generated content. However, uncertainty arises from heterogeneous sources, including epistemic and aleatoric components, so developing effective and well-grounded approaches to quantifying uncertainty remains an urgent and important issue~\cite{wang2025uncertainty,wang2025machinery,wang2025ascd,chen2024detecting}.

Our motivation stems from an intuitive observation: surface fluency alone is insufficient to adjudicate the reliability of model outputs in the presence of hallucinations, thereby necessitating the incorporation of Uncertainty Quantification (UQ) to assess answer credibility. Contemporary approaches typically extract salient generative signals---such as attention distributions, hidden-state dynamics, or token-level probability information---during decoding \cite{kadavath2022know,kuhn2023semantic}, use these signals to compute uncertainty scores at multiple granularities (e.g., token- and sentence-level) \cite{malinin2021uncertainty,lin2022trustincontext}, and subsequently aggregate them into a holistic credibility indicator. This indicator is then surfaced to end users via concise interface cues (e.g., ``high confidence'' or ``potentially unreliable''), providing real-time decision support in high-stakes domains such as medicine and psychology and thereby mitigating risks associated with hallucinated content.

Existing uncertainty estimation methods have demonstrated remarkable effectiveness in distinguishing correct from incorrect answers; however, such heuristic approaches fail to provide a provable risk guarantee (i.e., correctness coverage). To address this limitation, we introduce the framework of Conformal Prediction (CP)~\cite{angelopoulos2023conformal,peng2025conformal,barber2023conformal,wang2025coin,snell2025conformal,wang-etal-2025-sconu}. CP is a statistical learning paradigm designed to generate interpretable prediction sets with guaranteed confidence levels alongside model outputs. Its core principle lies in specifying a significance level (or allowable error rate), computing a \emph{nonconformity score} from historical data, and comparing it with the prediction of a new instance, thereby constructing a prediction set that, with high probability, contains the label. 

Compared to traditional probabilistic prediction, CP imposes minimal assumptions on data distribution, requiring only that samples satisfy the property of exchangeability—which endows it with strong adaptability and robustness across diverse tasks and model architectures. Moreover, the size of the prediction set can directly indicate the model's uncertainty: when confidence is low, the set expands to preserve coverage; when confidence is high, the set contracts to enhance decision precision. Owing to its provable coverage guarantee, CP is of particular importance in high-stakes applications such as medical diagnosis, legal text analysis, and scientific research, as it not only quantifies the reliability of predictions but also enables flexible trade-offs between risk and utility under varying confidence thresholds. In recent years, researchers have further extended CP beyond traditional classification and regression tasks to more complex domains such as open-ended natural language generation and structured prediction. By incorporating techniques such as sampling and reprompting, these advancements effectively address the challenges posed by vast output spaces and the inherent diversity of possible answers.

We undertake a systematic empirical investigation of six cutting-edge large language models—LLaMA-3.2-1B, Qwen2.5-3B-Instruct, Vicuna-7B-v1.5, Qwen2.5-7B-Instruct, LLaMA-3.1-8B-Instruct, and Vicuna-13B-v1.5—across the CoQA and TriviaQA benchmarks, with the objective of rigorously evaluating their uncertainty quantification capabilities. The assessment is grounded in two principled metrics: (1) Empirical Coverage Rate (ECR), quantifying the proportion of instances wherein the constructed prediction set encapsulates the ground-truth response, and (2) Average Prediction Set Size (APSS), measuring the expected cardinality of prediction sets as a proxy for epistemic uncertainty. Experimental analyses uncover pronounced disparities in ECR and APSS across models and datasets, underscoring the sensitivity of uncertainty estimates to both model-specific generative competence and dataset-inherent statistical properties. To address the inherent challenges of uncertainty estimation in open-ended generation, we introduce a token-level entropy-based nonconformity scoring paradigm within the conformal prediction framework. The proposed method enables the construction of calibrated prediction sets under minimal distributional assumptions and without reliance on labeled supervision. Empirical results substantiate the efficacy of the proposed approach in maintaining rigorous error control while achieving high coverage of ground-truth outputs, thereby furnishing robust reliability guarantees for model predictions in open-domain generative settings.

\section{Related Work}
\textbf{Uncertainty Analysis:}  Framed by uncertainty analysis, prior work traces a progression from general UQ to black-box UQ for LLMs and, ultimately, to CP-based coverage guarantees: in ML and NLP, UQ is foundational for decision-making and risk control, yet confidence-based metrics (e.g., entropy) are vulnerable to calibration mismatch, while Bayesian and ensemble paradigms—despite their theoretical appeal—incur prohibitive computational and engineering overheads at LLM scale. Early LLM efforts relied on white-box signals (token likelihoods, internal activations) and fine-tuning for improved calibration. However, API-level black-box constraints and resource budgets limit their practicality, motivating an output-only, black-box route: multi-sample generation is used to capture semantic consistency and dispersion, with the “most frequent generation” serving as a self-consistency anchor to construct actionable, internal-agnostic uncertainty surrogates. Building on this, conformal prediction (CP) offers a systematic bridge from heuristic scores to coverage-guaranteed prediction sets by using a small number of i.i.d. calibration instances to map external UQ scores to set thresholds and to achieve ground-truth coverage at a user-specified error rate, while remaining model- and distribution-agnostic and computationally feasible. Further, to accommodate the vast output space of open-ended generation, recent work extends CP from classification to sequence generation via stopping rules and sampling approximations, thereby making explicit both “when confidence is sufficient” and “how large the set  should be to remain robust”~\cite{iutzeler2025riskcontrolling}.

\textbf{Conformal Prediction:}  As a distribution-free statistical calibration framework, conformal prediction (CP) offers a systematic pathway from heuristic uncertainty scores to coverage-guaranteed prediction sets and has garnered increasing attention in ML and NLP (Angelopoulos and Bates, 2021; Lu et al., 2023). Its core mechanism, in the classification setting, excludes implausible labels via nonconformity scores to form a candidate set that, with high probability, contains the ground truth, while using the set size to quantify predictive uncertainty~\cite{wang2025sample}; this paradigm has been validated in part-of-speech tagging, paraphrase detection, and fact verification. Owing to its nonparametric, distribution-free, model-agnostic, and computationally efficient characteristics, CP is well suited to large language models; moreover, its ability to deliver distribution-free, finite-sample calibration of general risk functions provides a methodological bridge from discriminative tasks to generative ones. For open-ended text generation with an effectively unbounded output space, recent work extends CP from classification to sequences via sampling and stopping rules, making explicit both when confidence is sufficient and how large a set must be to remain robust—without enumerating the entire candidate space. By contrast, CP-style confidence intervals developed for diffusion-based image generation remain at the pixel level and do not directly transfer to the combinatorial and semantic structure of language, underscoring the need for CP frameworks tailored to free-form generation in LLMs.

\section{Method}
This section proposes a general framework grounded in semantic self-consistency and split conformal prediction to assess uncertainty in outputs generated by black-box language models and, on this basis, to construct prediction sets endowed with statistical coverage guarantees. Unlike traditional approaches that rely on log-likelihoods or internal distributions, we treat the language model as a complete black box: only standard API access to generated outputs is assumed, with no visibility into parameters, logits, or attention weights. Under this setting, the central challenge is to estimate confidence from a finite set of sampled outputs and to achieve a theoretically verifiable calibrated prediction.

\subsection{Problem Setup and Candidate Generation Mechanism}
Let \(x \in \mathcal{X}\) denote the input to a natural-language task (e.g., a question in QA or a passage in summarization), with reference answer \(y^* \in \mathcal{Y}\). For a black-box language model \(f(\cdot)\), we generate, for input \(x\), a set of candidate outputs \(\hat{\mathcal{Y}}(x)=\{\hat{y}_1,\hat{y}_2,\ldots,\hat{y}_M\}\), where \(M\) is a fixed number of generations. The decoding can employ various strategies (e.g., temperature sampling, beam search, or top-\(p\) sampling), which we abstract as
\begin{equation}
\hat{y}_m \sim f(x;\theta),\quad m=1,\ldots,M.
\end{equation}
Where \(\theta\) collects control hyperparameters (e.g., temperature, beam width) and is treated as fixed. Given the openness and diversity of natural-language generation, a single input \(x\) often admits multiple semantically equivalent yet lexically distinct outputs. We thus adopt a set-generation perspective, viewing all candidates as uncertain responses to \(x\) and seeking to extract information about output quality and confidence from their distributional structure. Our ultimate objective is to estimate the uncertainty \(U(\hat{y}_m)\) for each candidate \(\hat{y}_m\) and, at a user-specified error tolerance \(\alpha\), to construct a prediction set \(\Gamma(x)\) that satisfies the following coverage guarantee:
\begin{equation}
\mathbb{P}\big[y^* \in \Gamma(x)\big]\ge 1-\alpha.
\end{equation}
The construction is realized via quantile calibration in \S\ref{subsec:calibration}.

\begin{figure*}[!t]
    \centering
    \subfigure[Qwen2.5-3B-Instruct]{%
        \includegraphics[width=.32\textwidth]{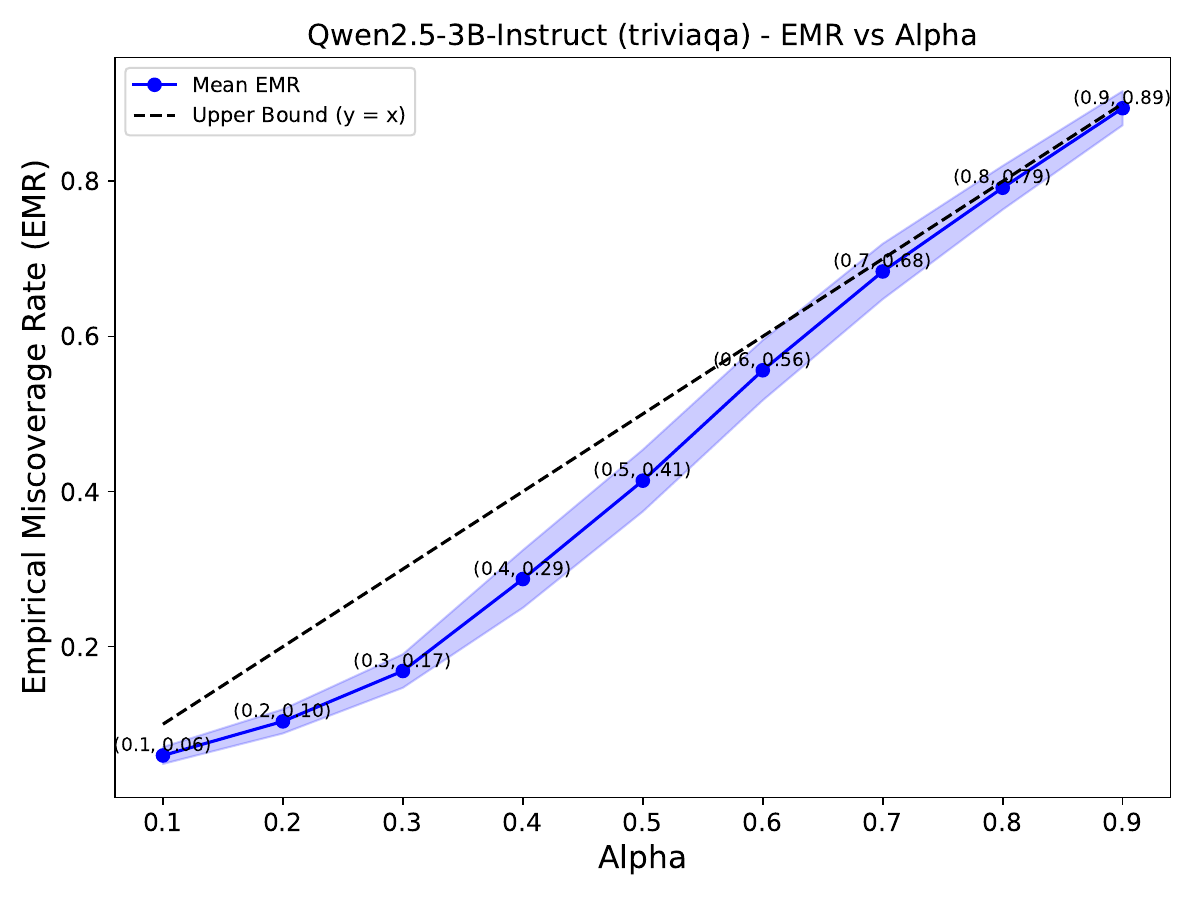}%
    }\hfill
    \subfigure[Llama-3.2-1B]{%
        \includegraphics[width=.32\textwidth]{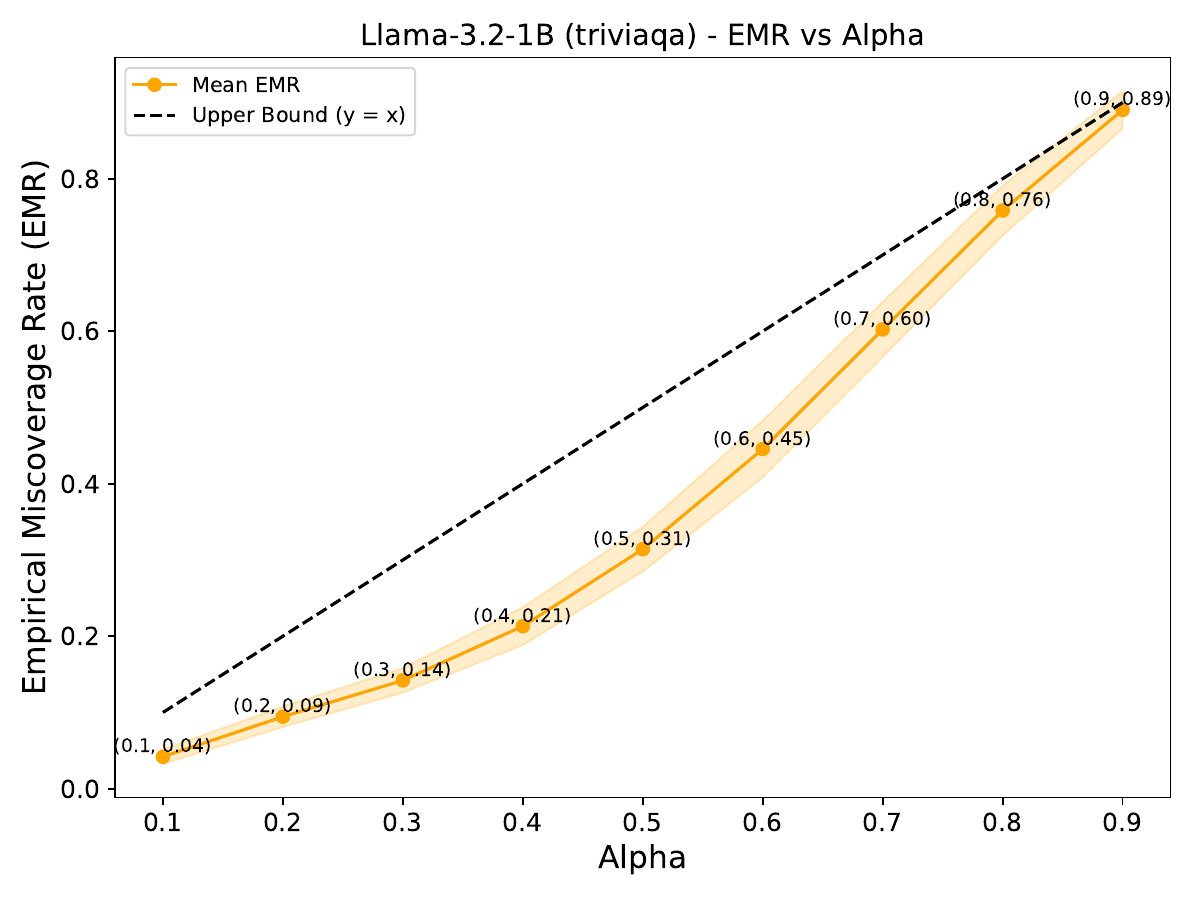}%
    }\hfill
    \subfigure[Qwen2.5-7B-Instruct]{%
        \includegraphics[width=.32\textwidth]{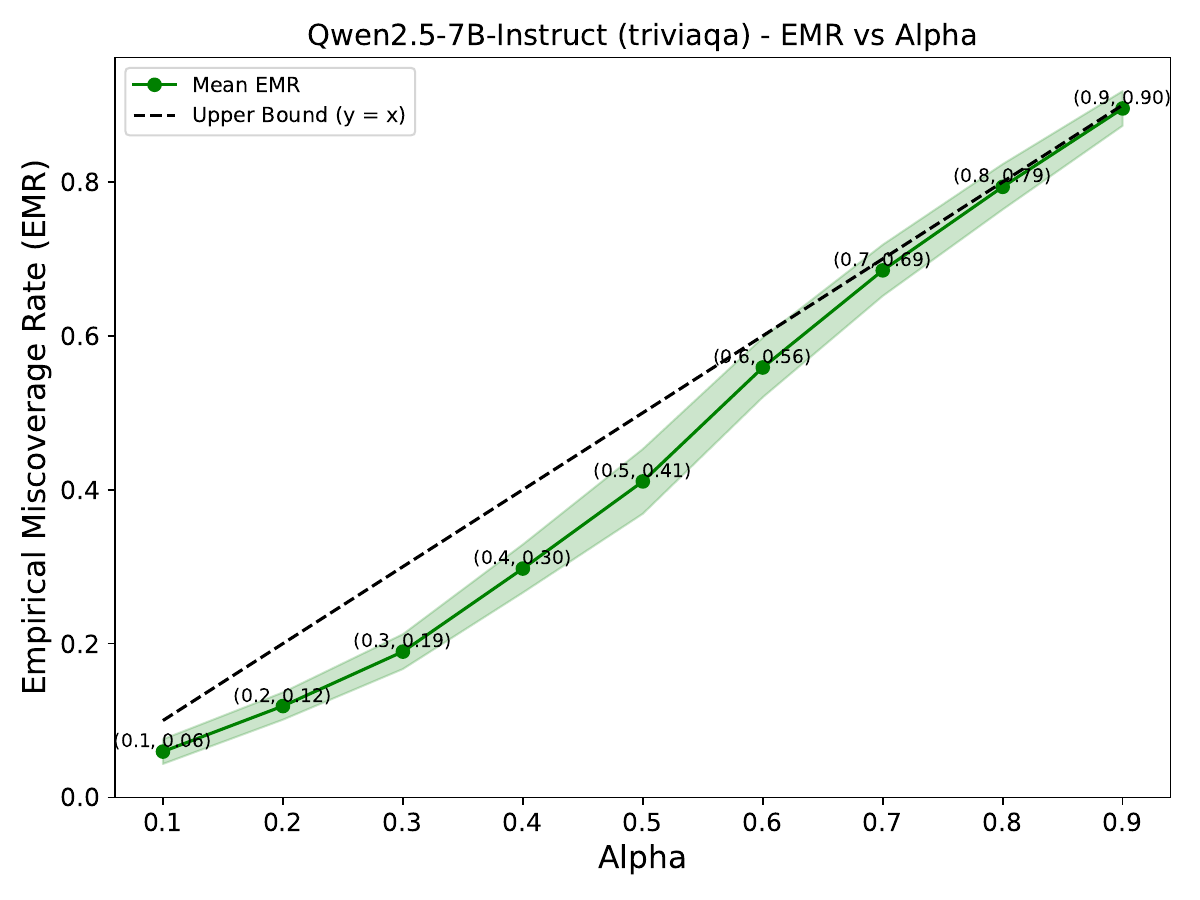}%
    }\\[-0.5em]
    \subfigure[Llama-3.1-8B-Instruct]{%
        \includegraphics[width=.32\textwidth]{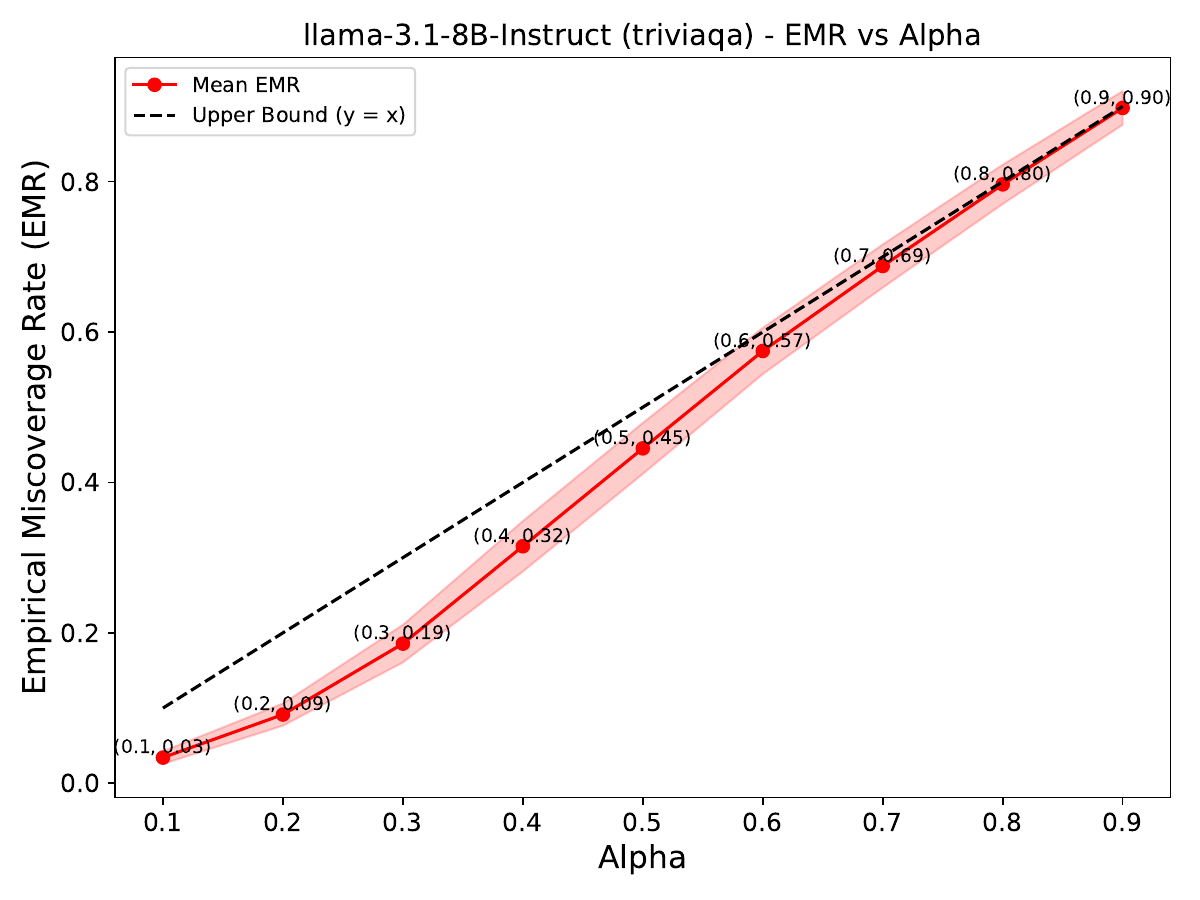}%
    }\hfill
    \subfigure[vicuna-7b-v1.5]{%
        \includegraphics[width=.32\textwidth]{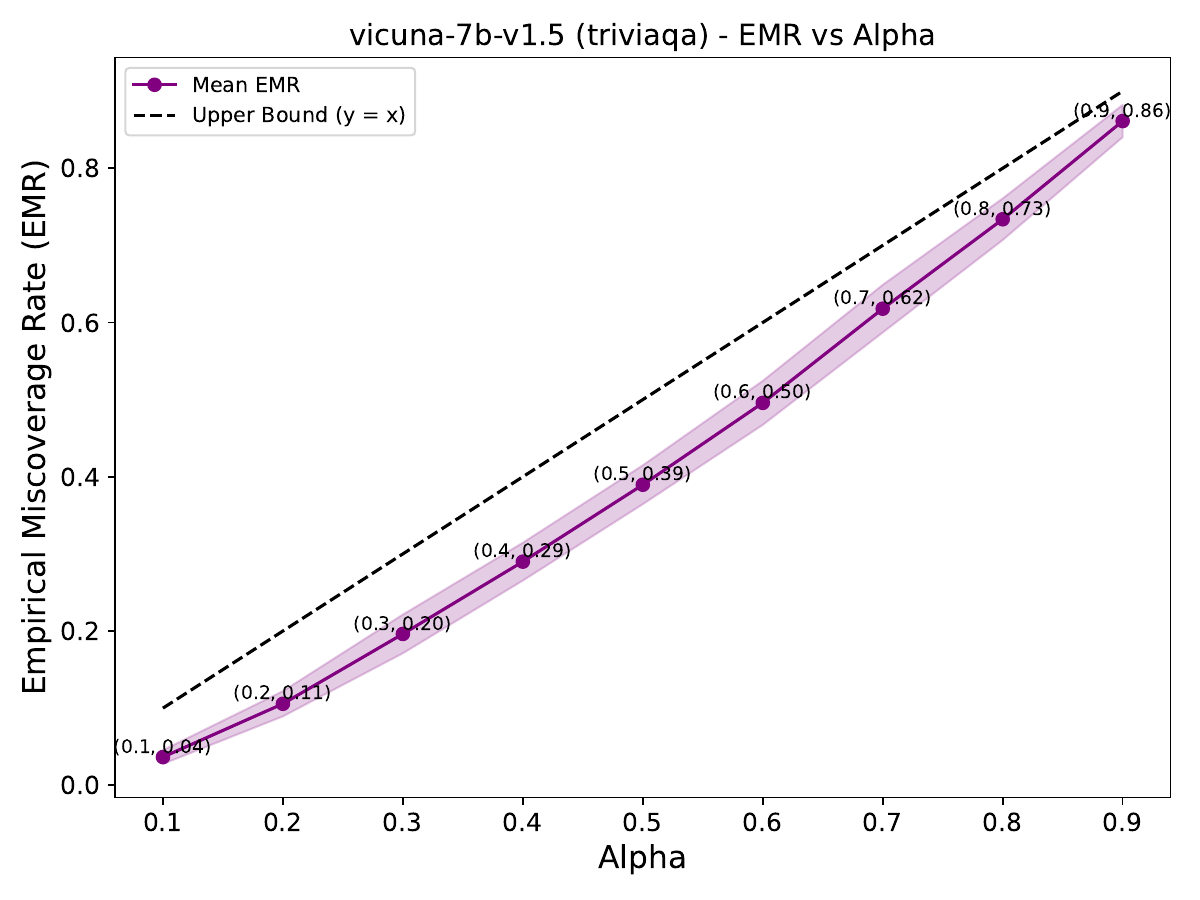}%
    }\hfill
    \subfigure[vicuna-13b-v1.5]{%
        \includegraphics[width=.32\textwidth]{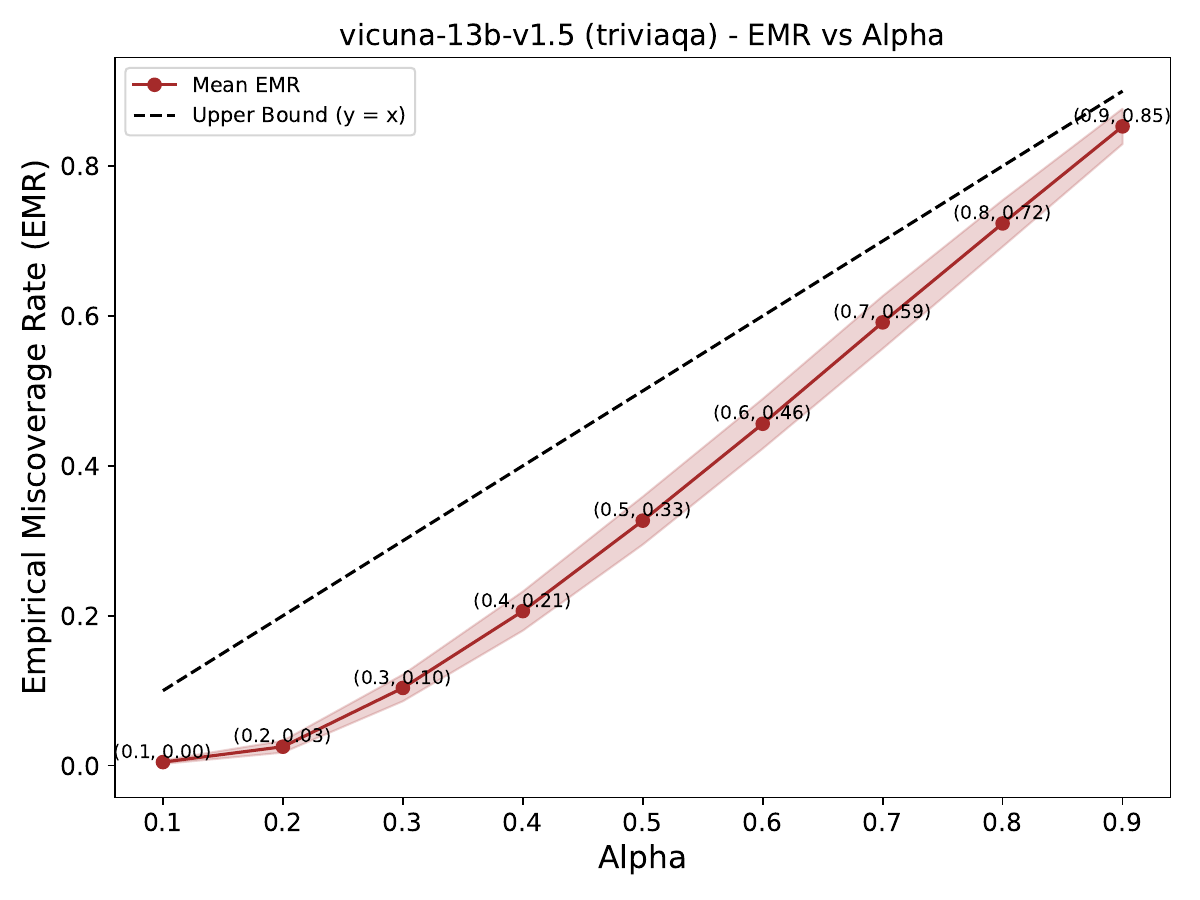}%
    }
    \caption{EMR vs. Alpha for six models on the TriviaQA dataset using \textbf{TECP}
}
\end{figure*}

\subsection{Uncertainty Estimation Driven by Inter-Candidate Semantic Consistency}
Because internal model probabilities are unavailable, we focus on the semantic agreement structure among generated texts and propose a semantic self-consistency--based scoring mechanism to define the uncertainty of each candidate. Specifically, for the candidate set \(\hat{\mathcal{Y}}(x)\) produced for input \(x\), we define the uncertainty \(U(\hat{y}_m)\) of candidate \(\hat{y}_m\) as a convex combination of two components: (i) the frequency with which other candidates are semantically equivalent to \(\hat{y}_m\), and (ii) the average semantic similarity between \(\hat{y}_m\) and all other candidates:
\begin{equation}
\begin{aligned}
U(\hat{y}_m)
&= 1 - \lambda \cdot
\underbrace{\frac{\big|\{\hat{y}_j \in \hat{\mathcal{Y}}(x):\, \hat{y}_j \approx \hat{y}_m\}\big|}{M}}_{\text{semantic-consistency frequency}} \\
&\quad - (1-\lambda)\cdot
\underbrace{\frac{1}{M}\sum_{j=1}^{M} S(\hat{y}_m,\hat{y}_j)}_{\text{average semantic similarity}}\!,
\end{aligned}
\end{equation}
Where \(S(\cdot,\cdot)\) denotes a sentence-level semantic similarity measure (e.g., BERTScore or cosine similarity), \(\hat{y}_j \approx \hat{y}_m\) denotes semantic equivalence, and \(\lambda\in[0,1]\) is a trade-off parameter. Intuitively, candidates that align semantically with the majority exhibit lower uncertainty, whereas those that deviate from the dominant semantics are assigned higher uncertainty. In practice, to avoid reliance on potentially costly clustering, we employ \emph{token-level average entropy} as a proxy for uncertainty, defined by
\begin{equation}
\begin{aligned}
U(\hat{y}_m)
&= \sum_{t=1}^{L_m} H_t \\
&= -\sum_{t=1}^{L_m}\sum_{v\in \mathcal{V}} p_t(v)\,\log p_t(v)\,,
\end{aligned}
\end{equation}
Where \(L_m\) is the length of \(\hat{y}_m\), \(\mathcal{V}\) is the vocabulary, and \(p_t(v)\) is the model’s predictive distribution over tokens \(v\) at position \(t\). Higher cumulative entropy indicates greater uncertainty and, hence, lower confidence in the generation. We adopt token-level entropy as our uncertainty metric because it captures the model's intrinsic uncertainty during generation, avoiding the overconfidence bias and hallucination sensitivity inherent in frequency-based diversity measures.This reference-free criterion evaluates each candidate’s confidence independently of the ground truth.

\subsection{Uncertainty Calibration and Prediction-Set Construction}
\label{subsec:calibration}

Given an uncertainty score for each candidate, we adopt the Split Conformal Prediction (SCP) framework: the available data are partitioned into a \emph{calibration set} and a \emph{test set}; the former is used to estimate a confidence threshold, which is then applied to the latter to build prediction sets with formal guarantees.
\begin{enumerate}
\item \textbf{Data partitioning and filtering.} For each sample \((x_i, y_i^*)\), if there exists a candidate \(\hat{y}_m\) such that \(S(\hat{y}_m, y_i^*)\ge \tau\), the sample is deemed \emph{assessable} (i.e., of acceptable generation quality) Moreover, retained in \(\mathcal{D}_{\text{filtered}}\), where \(\tau\) is a semantic-matching threshold typically in the range \(0.8\)–\(0.9\). We then randomly split \(\mathcal{D}_{\text{filtered}}\) into a calibration subset \(\mathcal{D}_{\text{cal}}\) and a test subset \(\mathcal{D}_{\text{test}}\) in a user-specified ratio.
\item \textbf{Construction of the nonconformity score multiset \(\mathcal{R}\).} On \(\mathcal{D}_{\text{cal}}\), collect uncertainty scores of the (semantically correct) candidates:
\begin{equation}
\begin{aligned}
\mathcal{R}
= \big\{\, U\big(\hat{y}^{(i)}_m\big)\ \big|\ &(x_i,y_i^*)\in \mathcal{D}_{\text{cal}},\\
&\hat{y}^{(i)}_m \text{ is semantically correct}\,\big\}\!.
\end{aligned}
\end{equation}
Using all candidates is a permissible, more conservative variant.
\item \textbf{Quantile-based confidence threshold.} For target coverage \(1-\alpha\), order \(\mathcal{R}\) increasingly and take the \(\lceil (1-\alpha)(n+1)\rceil\)-th order statistic as the empirical threshold \(\hat{q}_\alpha\):
\begin{equation}
\begin{aligned}
\hat{q}_\alpha
&= \mathrm{Quantile}\!\big(\mathcal{R},\, q_{\text{level}}\big),\\
q_{\text{level}}
&= \frac{\lceil(1-\alpha)(n+1)\rceil}{n}\,,
\end{aligned}
\end{equation}
Where \(n=|\mathcal{R}|\). We use a ``higher'' interpolation rule to avoid underestimating the threshold and thereby preserve coverage.
\item \textbf{Prediction-set construction.} For any test input \(x\), define
\begin{equation}
\begin{aligned}
\Gamma(x)
&= \big\{\, \hat{y}_m \in \hat{\mathcal{Y}}(x)\ :\ \\
&\quad U(\hat{y}_m)\le \hat{q}_\alpha \,\big\}\,.
\end{aligned}
\end{equation}
By construction, \(\Gamma(x)\) achieves a nominal coverage of at least \(1-\alpha\) for the ground-truth answer under standard exchangeability assumptions.
\end{enumerate}

\begin{figure*}[!t]
    \centering
    \subfigure[Qwen2.5-3B-Instruct]{%
        \includegraphics[width=.32\textwidth]{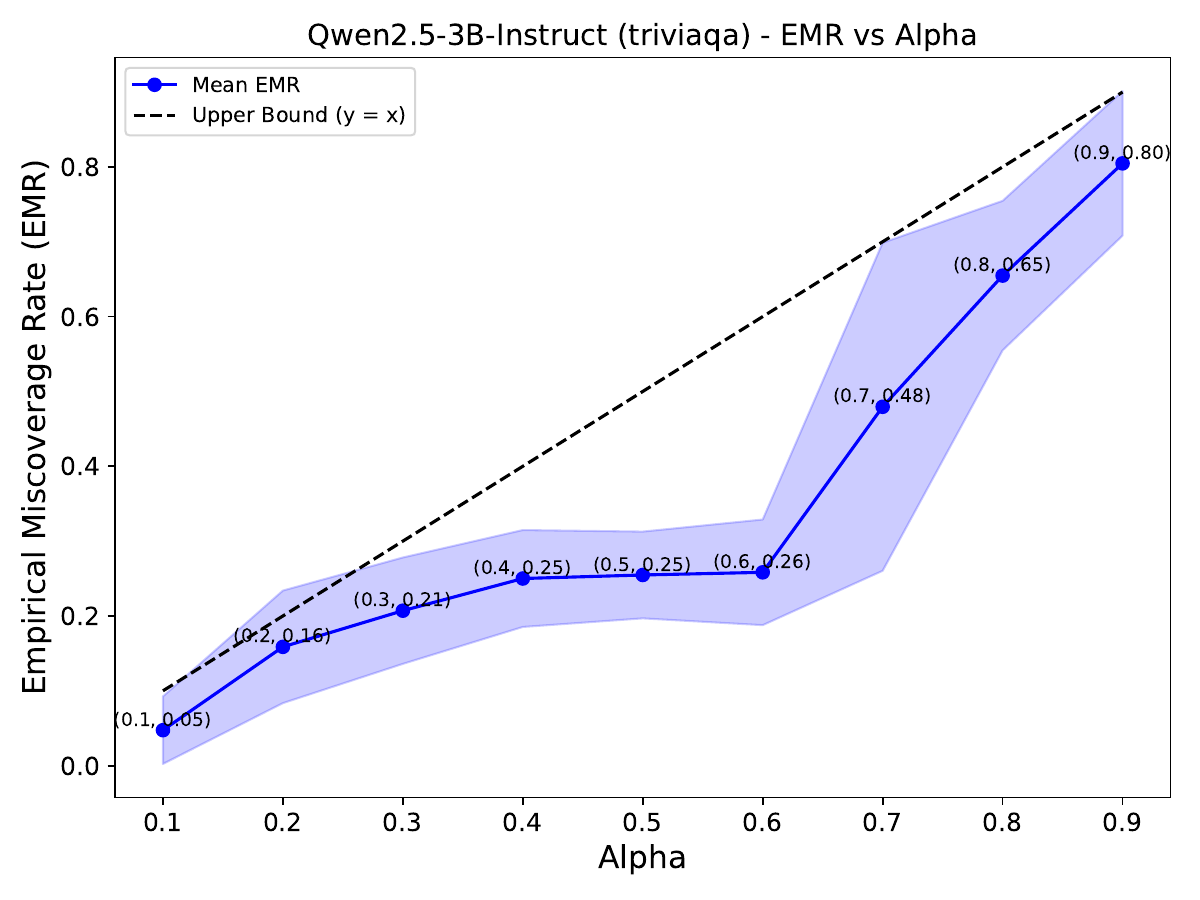}%
    }\hfill
    \subfigure[Llama-3.2-1B]{%
        \includegraphics[width=.32\textwidth]{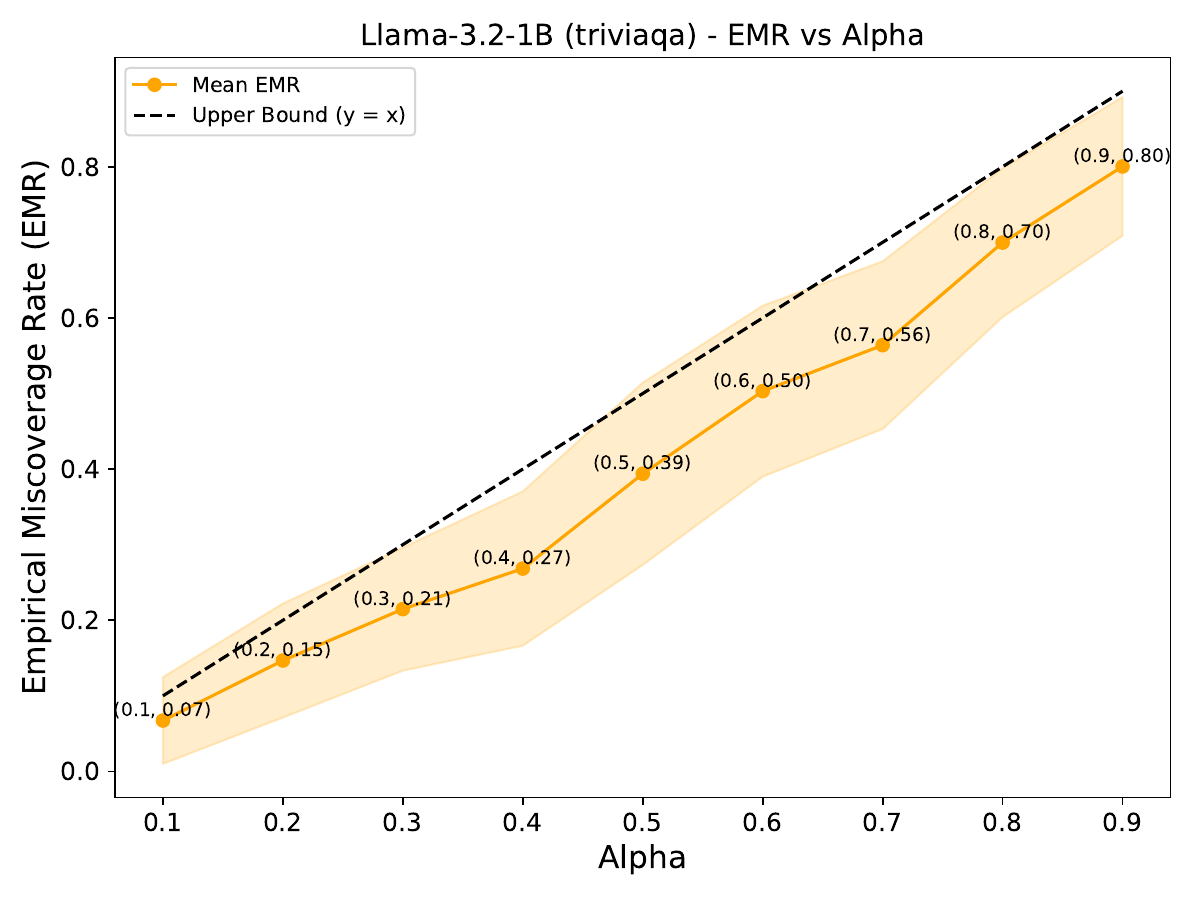}%
    }\hfill
    \subfigure[Qwen2.5-7B-Instruct]{%
        \includegraphics[width=.32\textwidth]{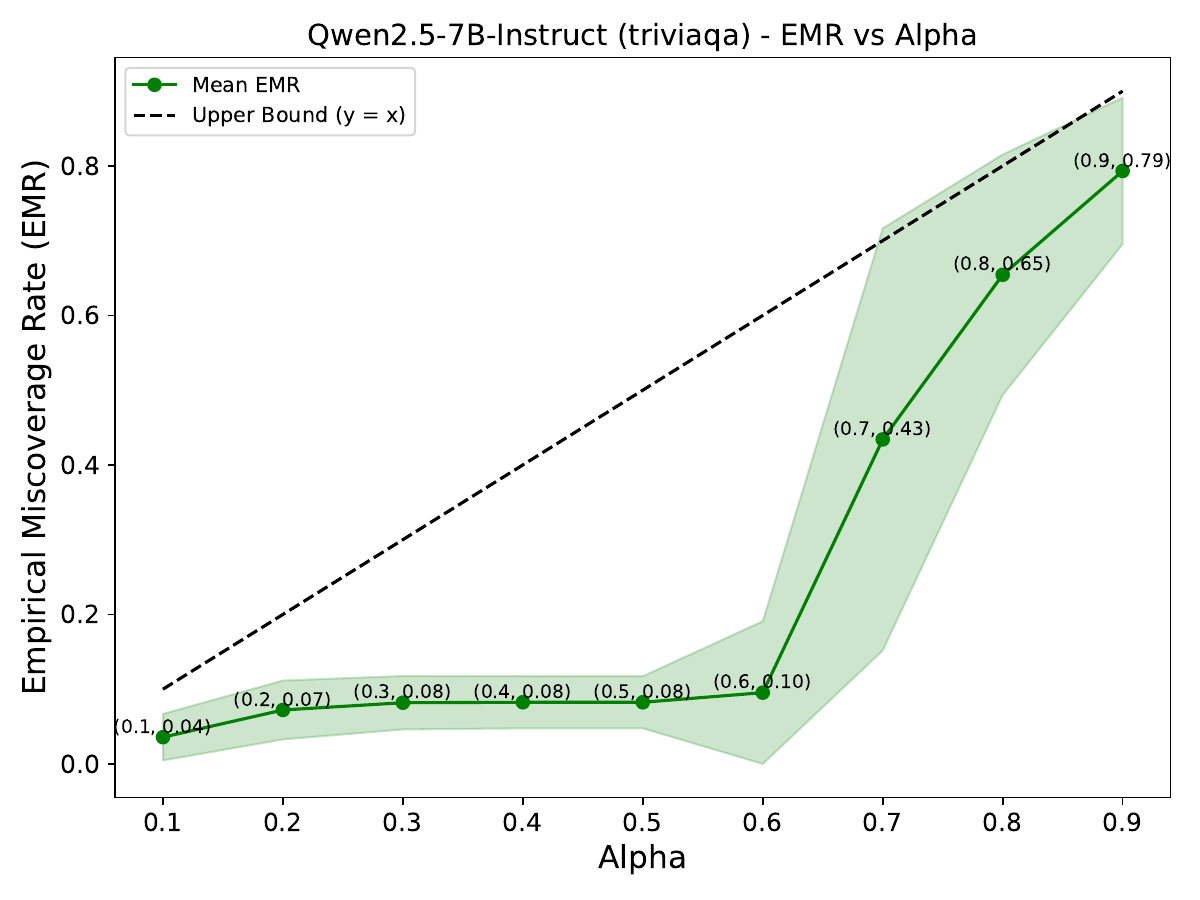}%
    }\\[-0.5em]
    \subfigure[Llama-3.1-8B-Instruct]{%
        \includegraphics[width=.32\textwidth]{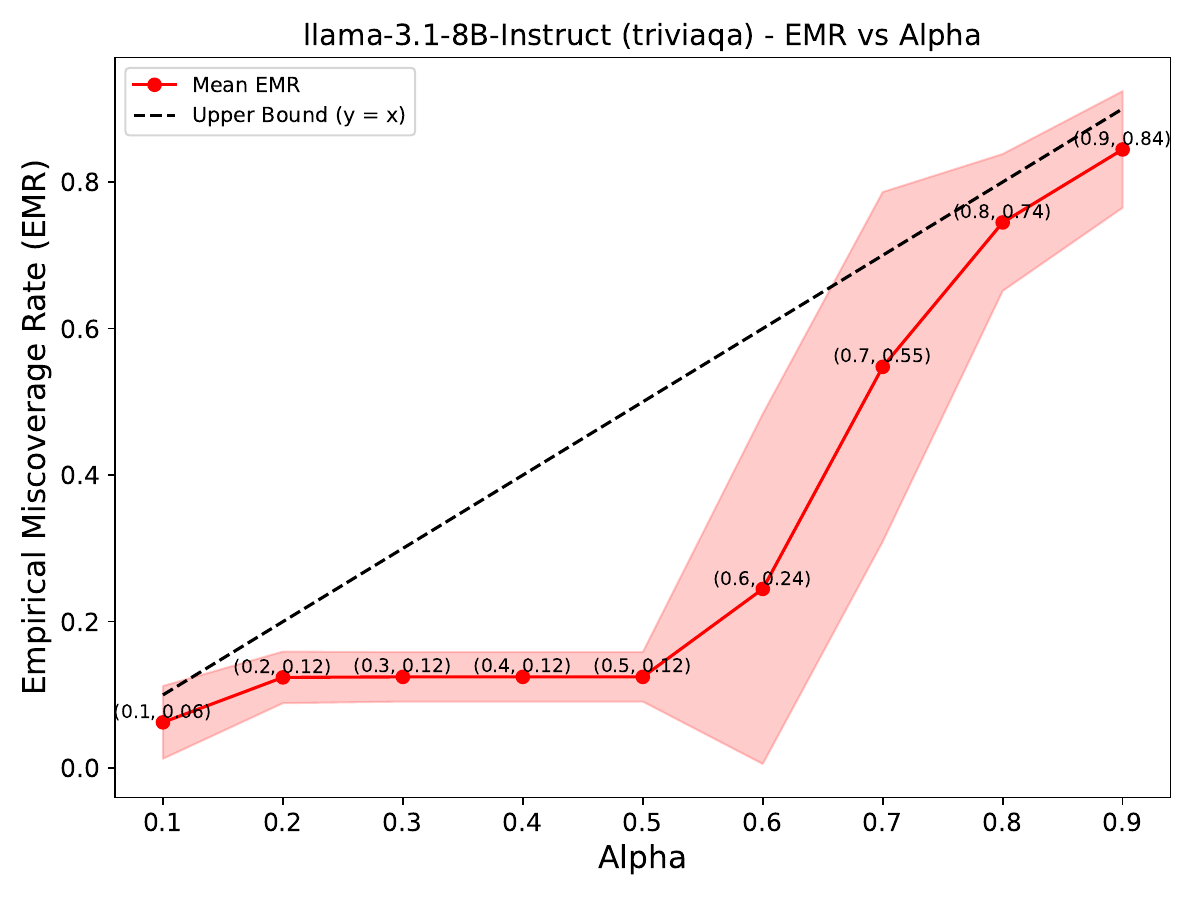}%
    }\hfill
    \subfigure[vicuna-7b-v1.5]{%
        \includegraphics[width=.32\textwidth]{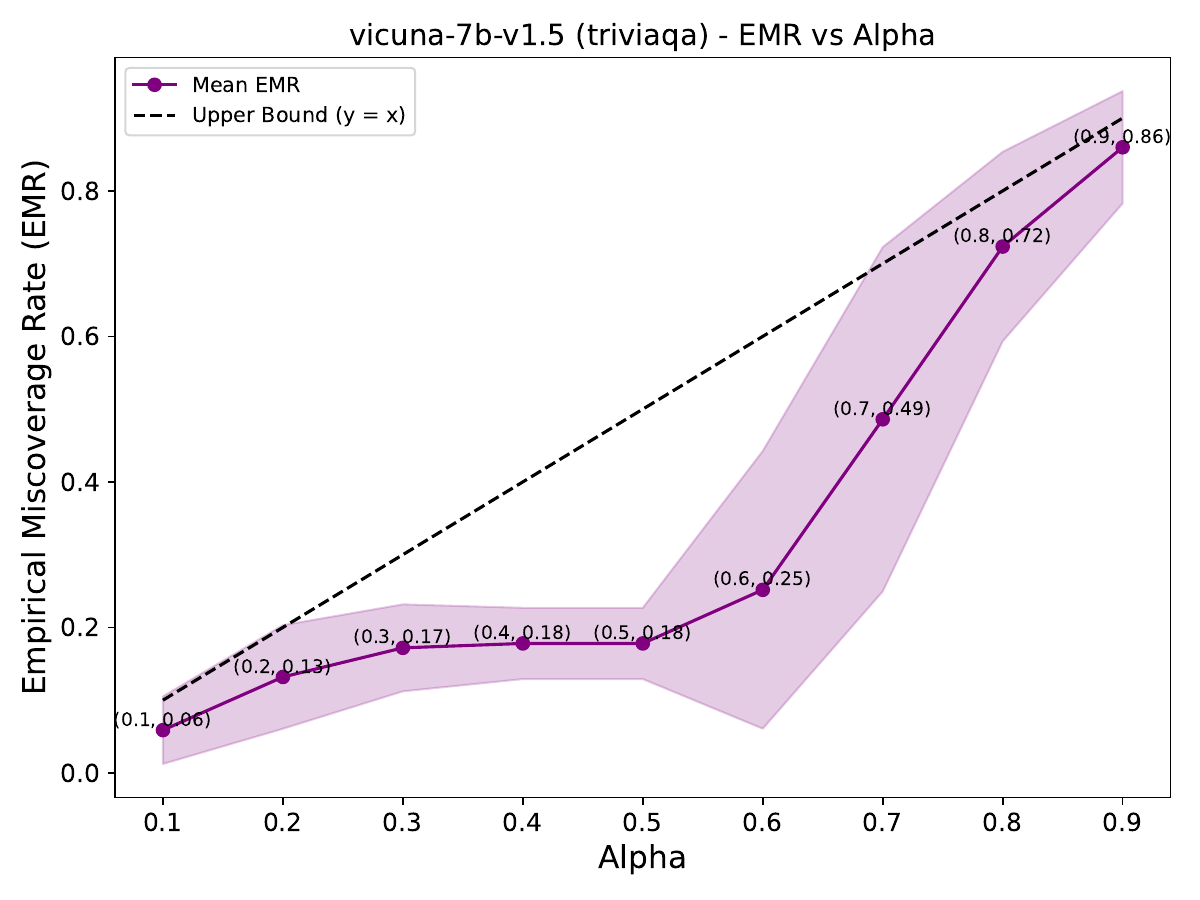}%
    }\hfill
    \subfigure[vicuna-13b-v1.5]{%
        \includegraphics[width=.32\textwidth]{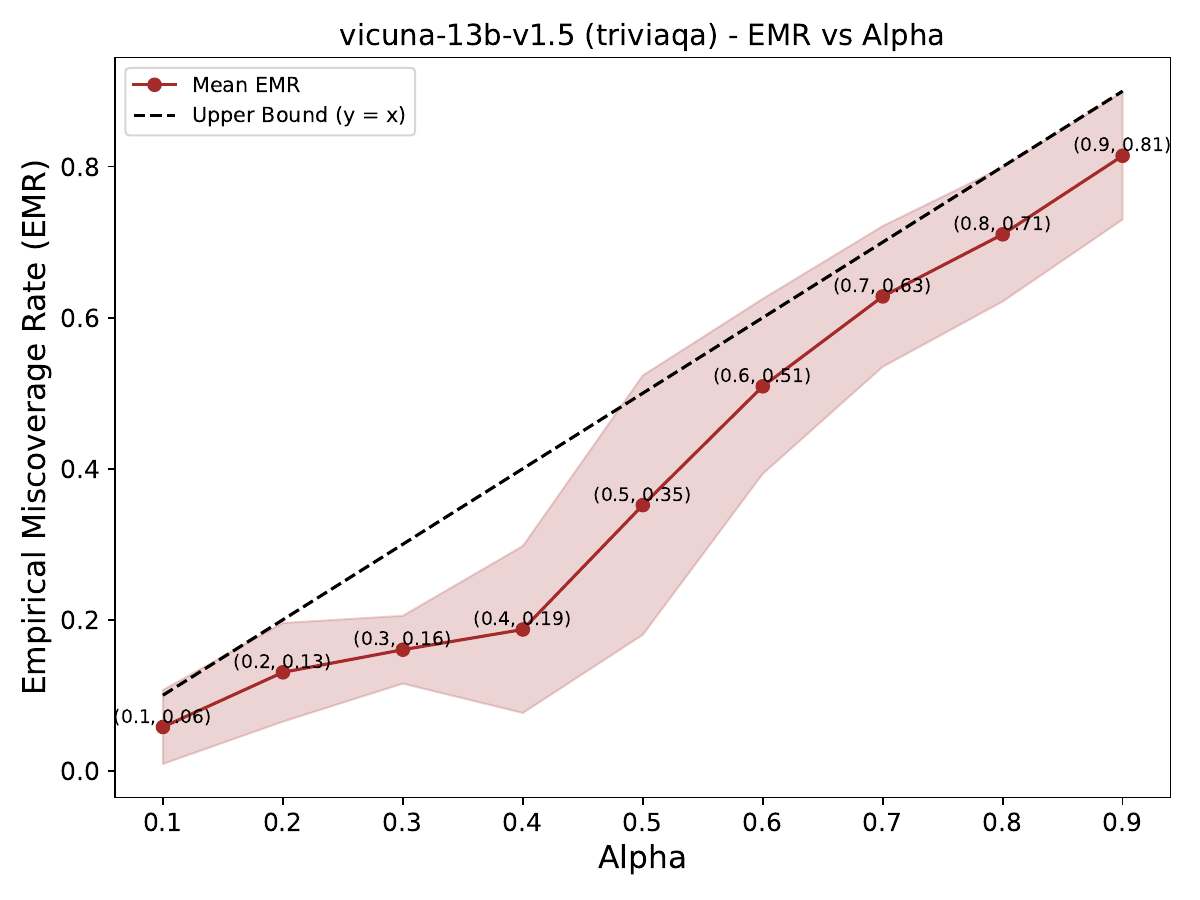}%
    }
    \caption{EMR vs. Alpha for six models on the TriviaQA dataset using \textbf{ConU}
}
\end{figure*}

\section{Experiments}
\subsection{Experimental Setup}
\textbf{Backbone LLMs and Evaluation Tasks.}  Since we incorporate token entropy into the conformal prediction framework for open-ended Question Answering (QA) tasks, it is necessary to evaluate the performance of the proposed architecture rigorously. To this end, we select a diverse set of open-source large language models (LLMs) as backbone models, including the \textbf{Llama series} (Llama-3.2-1B, Llama-3.1-8B-Instruct), the \textbf{Qwen series} (Qwen2.5-3B-Instruct, Qwen2.5-7B-Instruct), and the \textbf{Vicuna series }(Vicuna-7B-v1.5, Vicuna-13B-v1.5).

\textbf{Datasets.} We employ \textbf{TriviaQA} and \textbf{CoQA} as our experimental datasets. The \textbf{TriviaQA} dataset contains over 650,000 question–answer–evidence triples, covering a wide range of topics such as history, science, and entertainment, and supports multiple tasks including open-domain Question Answering and extractive Question Answering. \textbf{CoQA}, The dataset is large-scale and intended to facilitate the development and construction of conversational Question Answering systems.

\begin{figure*}[!t]
    \centering
    \subfigure[Qwen2.5-3B-Instruct]{%
        \includegraphics[width=.32\textwidth]{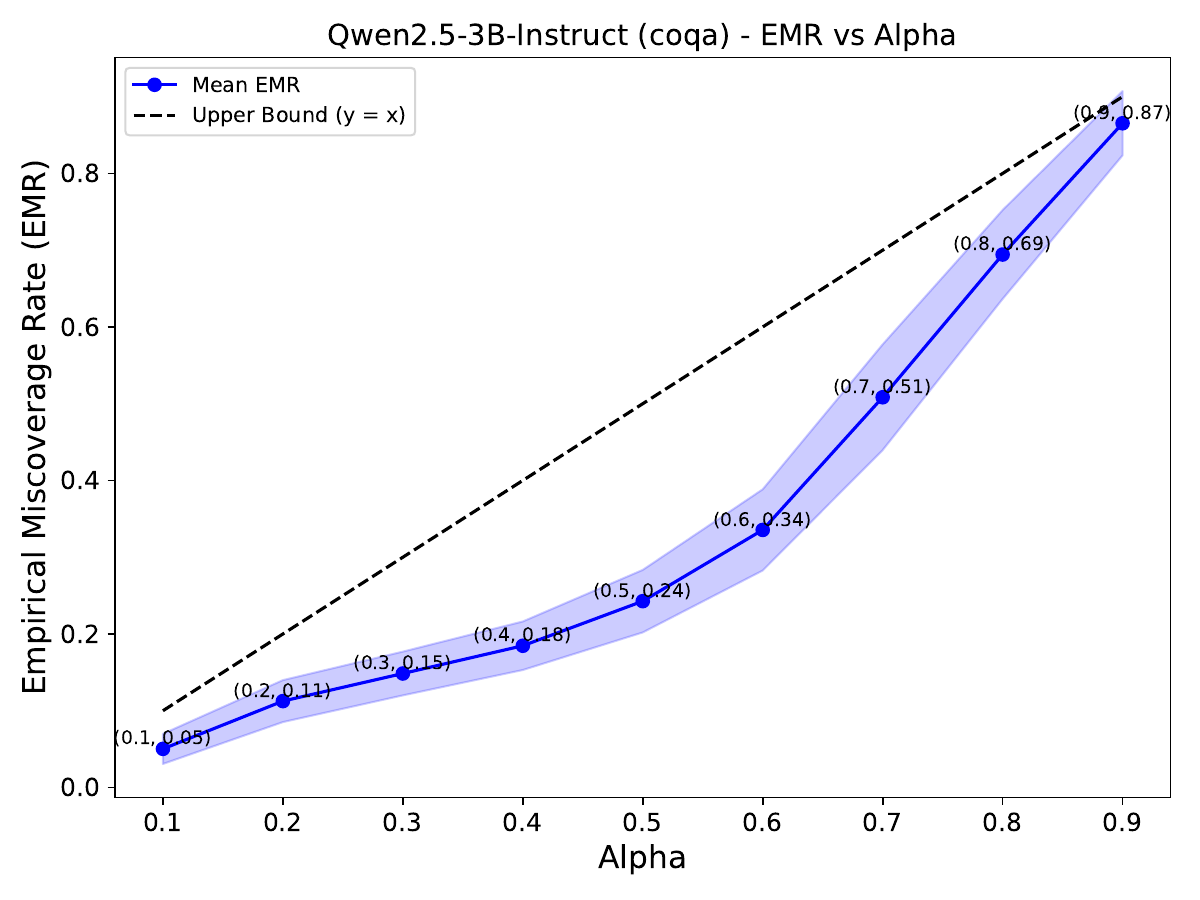}%
    }\hfill
    \subfigure[Llama-3.2-1B]{%
        \includegraphics[width=.32\textwidth]{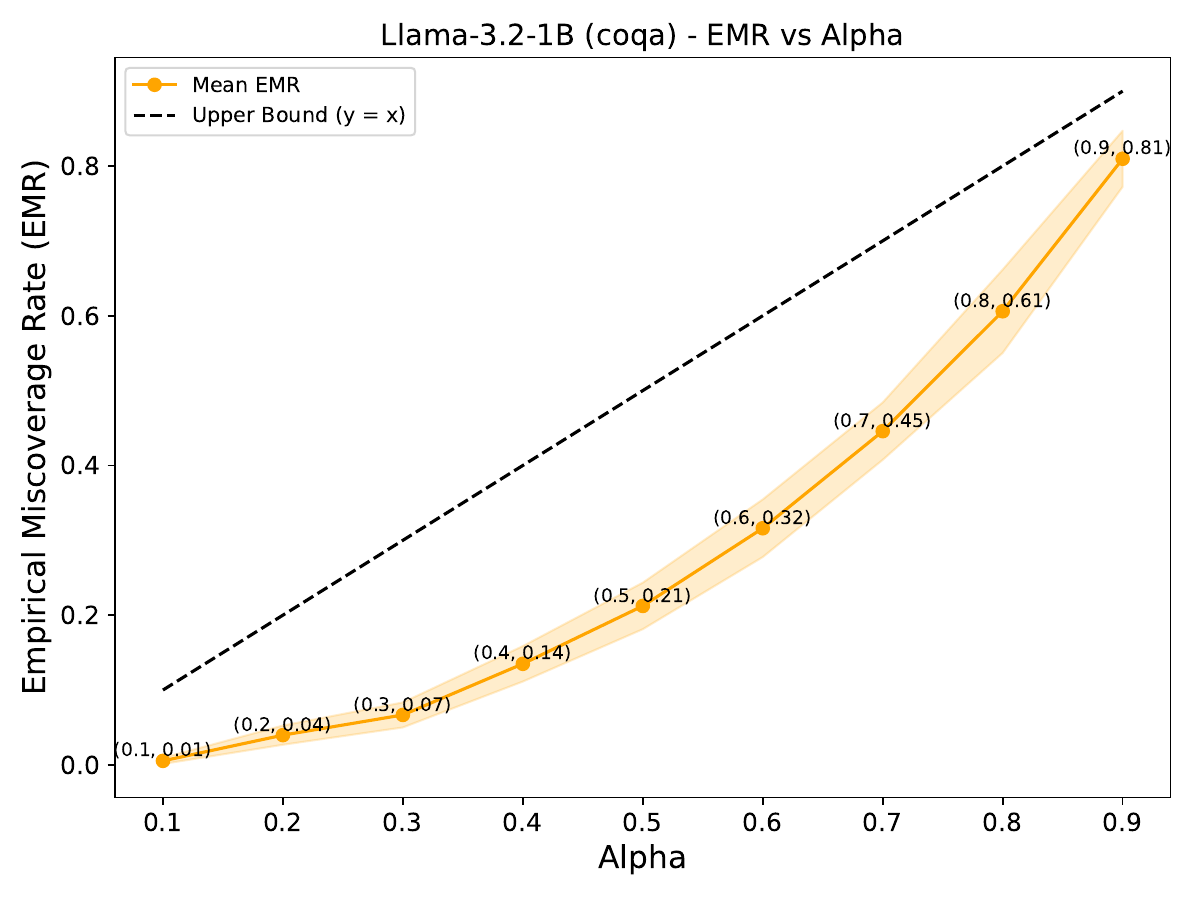}%
    }\hfill
    \subfigure[Qwen2.5-7B-Instruct]{%
        \includegraphics[width=.32\textwidth]{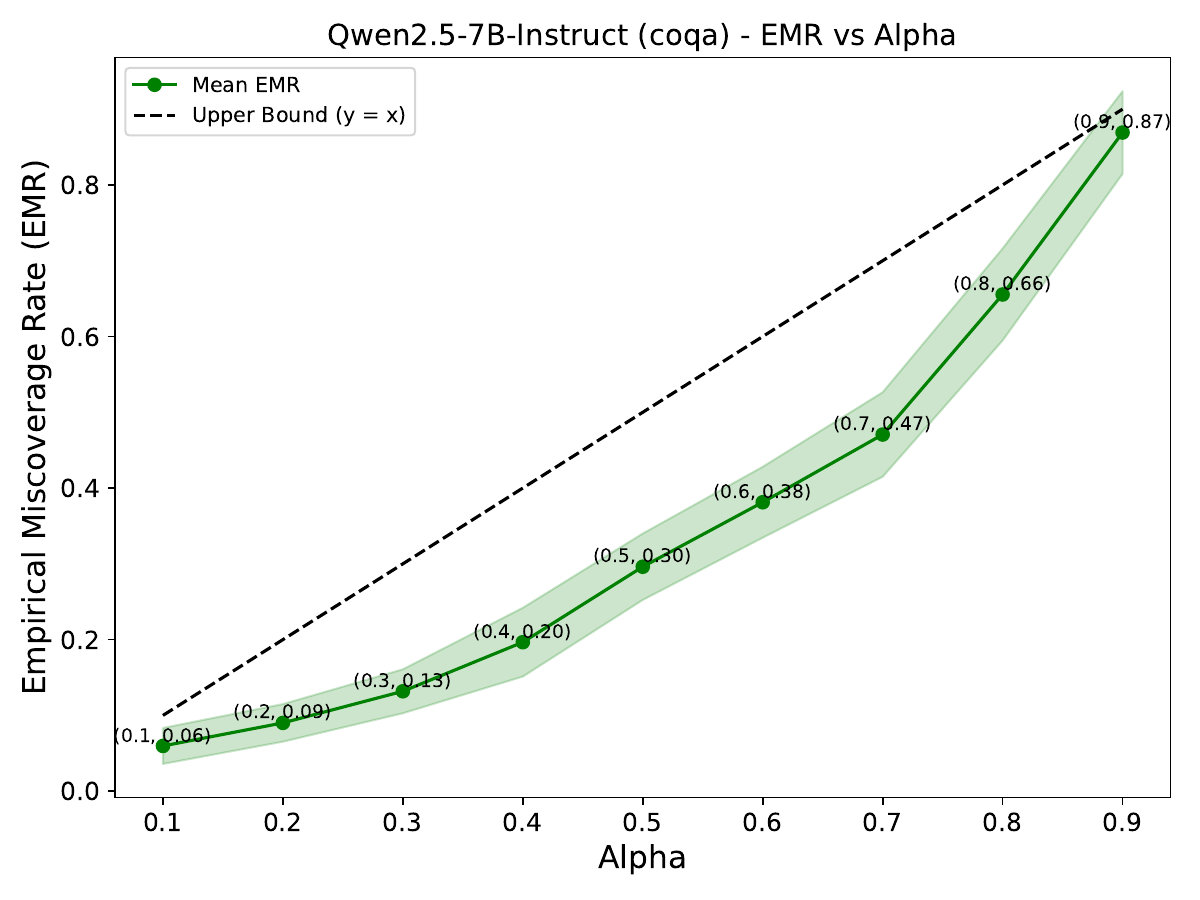}%
    }\\[-0.5em]
    \subfigure[Llama-3.1-8B-Instruct]{%
        \includegraphics[width=.32\textwidth]{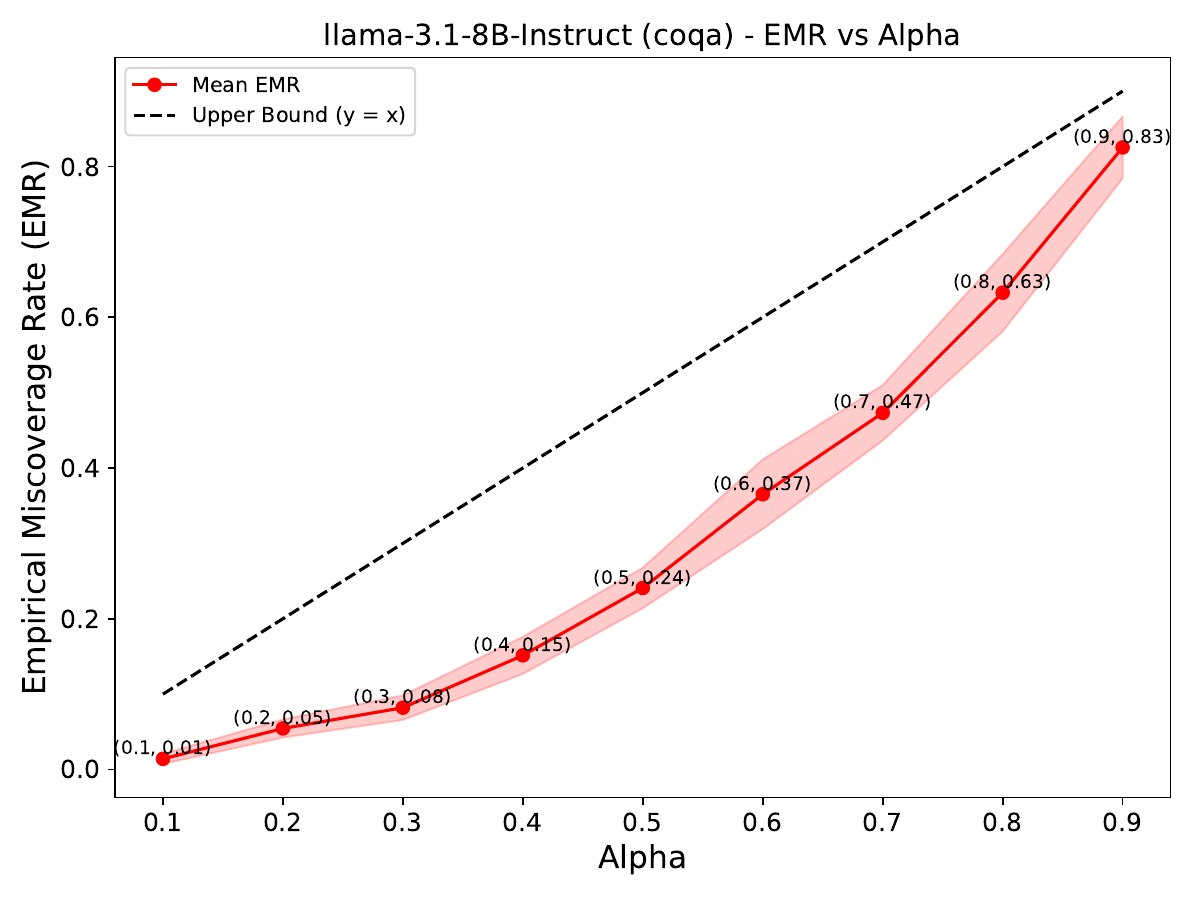}%
    }\hfill
    \subfigure[vicuna-7b-v1.5]{%
        \includegraphics[width=.32\textwidth]{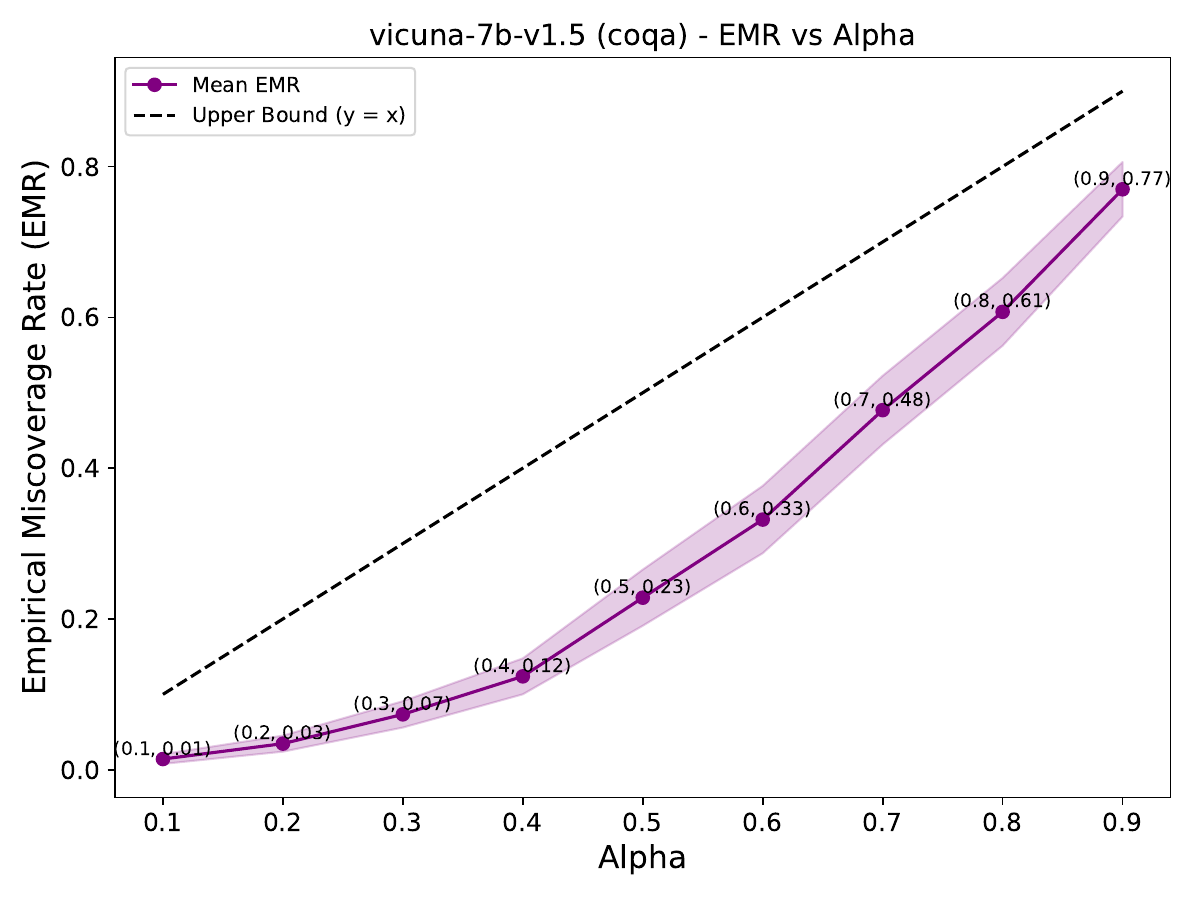}%
    }\hfill
    \subfigure[vicuna-13b-v1.5]{%
        \includegraphics[width=.32\textwidth]{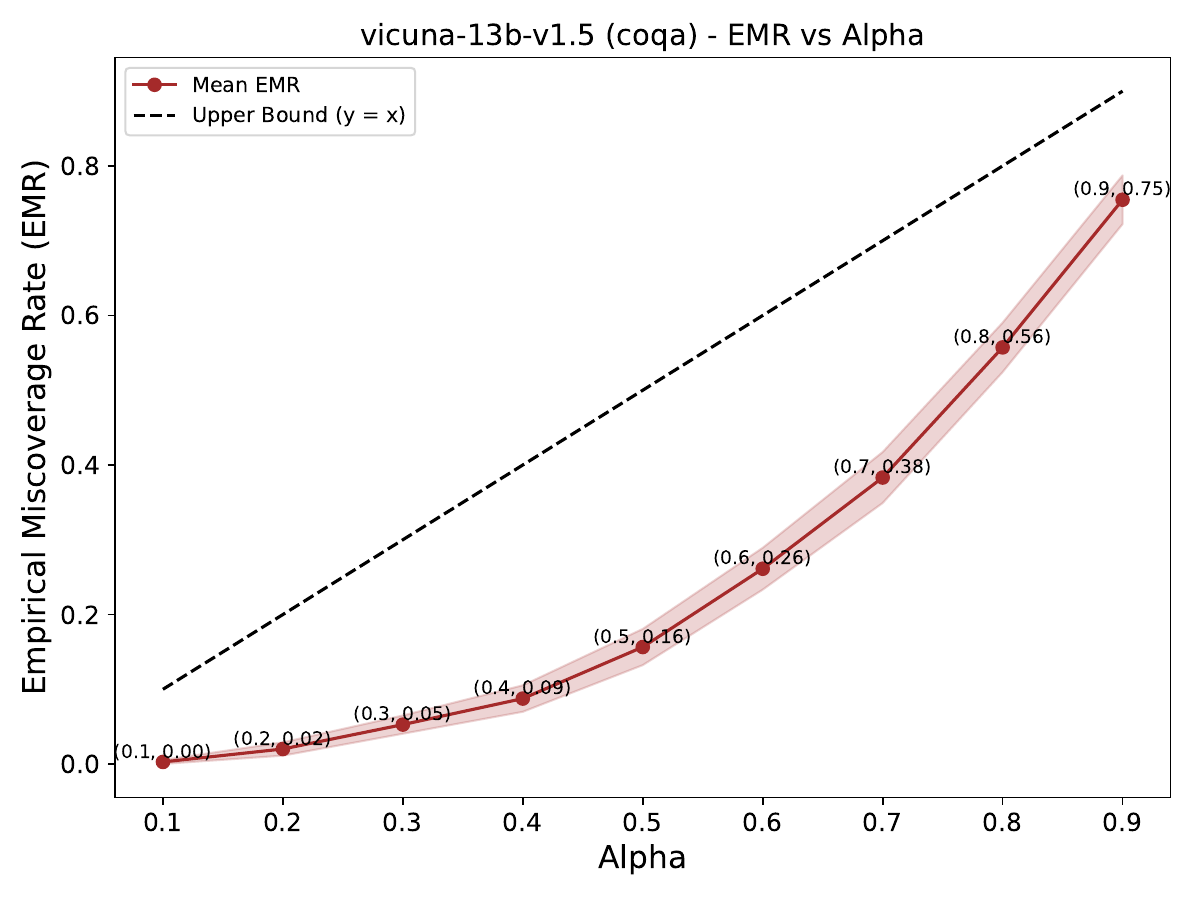}%
    }
    \caption{EMR vs. Alpha for six models on the CoQA dataset using \textbf{TECP}}
\end{figure*}

\textbf{Baseline.} The baseline method adopts a \textit{predictive uncertainty}-based strategy, which constructs the prediction set by quantifying the model's uncertainty over its generated outputs:
\vspace{-3mm}
\begin{itemize}
  \item \textbf{ConU.} A prediction method is proposed based on \textbf{Conformal Uncertainty}, which evaluates candidate responses generated for each input prompt and quantifies predictive uncertainty by incorporating both model confidence scores and calibration error. In contrast to heuristic approaches that rely solely on response-level variability, ConU offers a principled mechanism for constructing prediction sets with \textbf{theoretical risk control guarantees}. Notably, the method does not require access to model log-probabilities, yet is still able to retain high-confidence responses selectively. It enables the construction of prediction sets that are statistically guaranteed to meet predefined risk levels, while remaining representative, stable, and interpretable.

\end{itemize}

\textbf{Metrics.} We use the following metrics for evaluation:
\vspace{-3mm}
\begin{itemize}
  \item \textbf{Expected Metric Recall (EMR)} quantifies the proportion of instances where the prediction set contains at least one output exceeding a predefined performance threshold. It captures the set's informativeness.
  \vspace{-3mm}
  \item \textbf{Average Prediction Set Size (APSS)} computes the mean cardinality of prediction sets across the evaluation corpus. Lower values indicate greater efficiency.
\end{itemize}

\textbf{Correctness Evaluation.} We adopt a similarity-based criterion to evaluate the alignment between model predictions and ground-truth answers. Each generated response is paired with its corresponding reference answer and assessed using a semantic matching model. To this end, we leverage a DistillRoBERTa-based cross-encoder, which scores the semantic closeness between the two texts. A prediction is deemed correct if its similarity score surpasses a fixed threshold of 0.7, ensuring only semantically faithful outputs are retained. This approach enables a fine-grained evaluation of correctness beyond surface-level token overlap.

\subsection{Results for QA}
Systematic experiments are conducted on two open-domain question answering datasets, TriviaQA and CoQA, to evaluate the coverage performance and prediction set size of the proposed method across different language models. Figure 1 reports the EMR results on TriviaQA for six models: Qwen2.5-3B-Instruct, Qwen2.5-7B-Instruct, LLaMA-3.2-1B, LLaMA-3.1-8B-Instruct, Vicuna-7B-v1.5, and Vicuna-13B-v1.5; while Figure 3 presents the corresponding results on CoQA. All experiments adopt a sampling size of 10, where predictions are ranked according to predictive uncertainty and calibrated through the conformal prediction framework to construct output sets with formal coverage guarantees. At a risk level of $\alpha < 0.2$, EMR remains consistently below 0.1 for all models except the relatively weaker Vicuna-7B-v1.5 (EMR = 0.11), indicating that the method can generate reliable and high-confidence prediction sets under risk-controlled settings. Moreover, as shown in Figures 1, EMR tends to decrease as model capacity improves (e.g., from Qwen2.5-3B to Qwen2.5-7B to Vicuna-13B), suggesting that stronger models yield more concentrated and selective prediction sets under the same uncertainty-based ranking and conformal calibration strategy.

Meanwhile, the APSS results (Table 1) show that the size of the prediction sets decreases monotonically as the risk level increases: at $alpha=0.1$, the average set size is close to 9 candidates, while at $alpha=0.9$ it contracts to about 1 candidate. This trend remains consistent across both TriviaQA and CoQA, with only minor differences among models, underscoring the stability of the method and its cross-task generalizability. Notably, larger-scale models exhibit a faster reduction in set size at medium to high risk levels, indicating that their prediction sets are more selective. In addition, the prediction set sizes are highly similar across all models in Table 1, suggesting that conformal prediction is model-agnostic.

\begin{figure*}[!t]
    \centering
    \subfigure[Qwen2.5-3B-Instruct]{%
        \includegraphics[width=.32\textwidth]{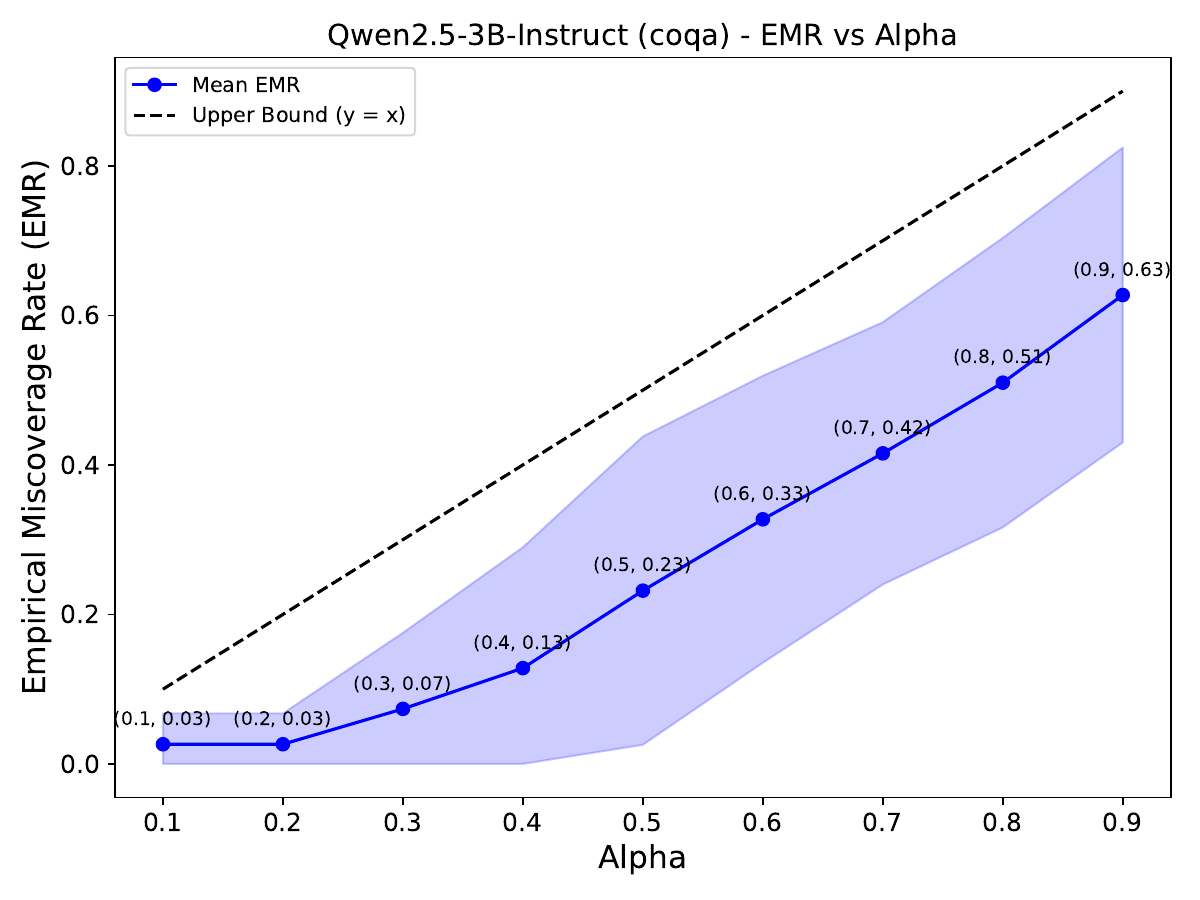}%
    }\hfill
    \subfigure[Llama-3.2-1B]{%
        \includegraphics[width=.32\textwidth]{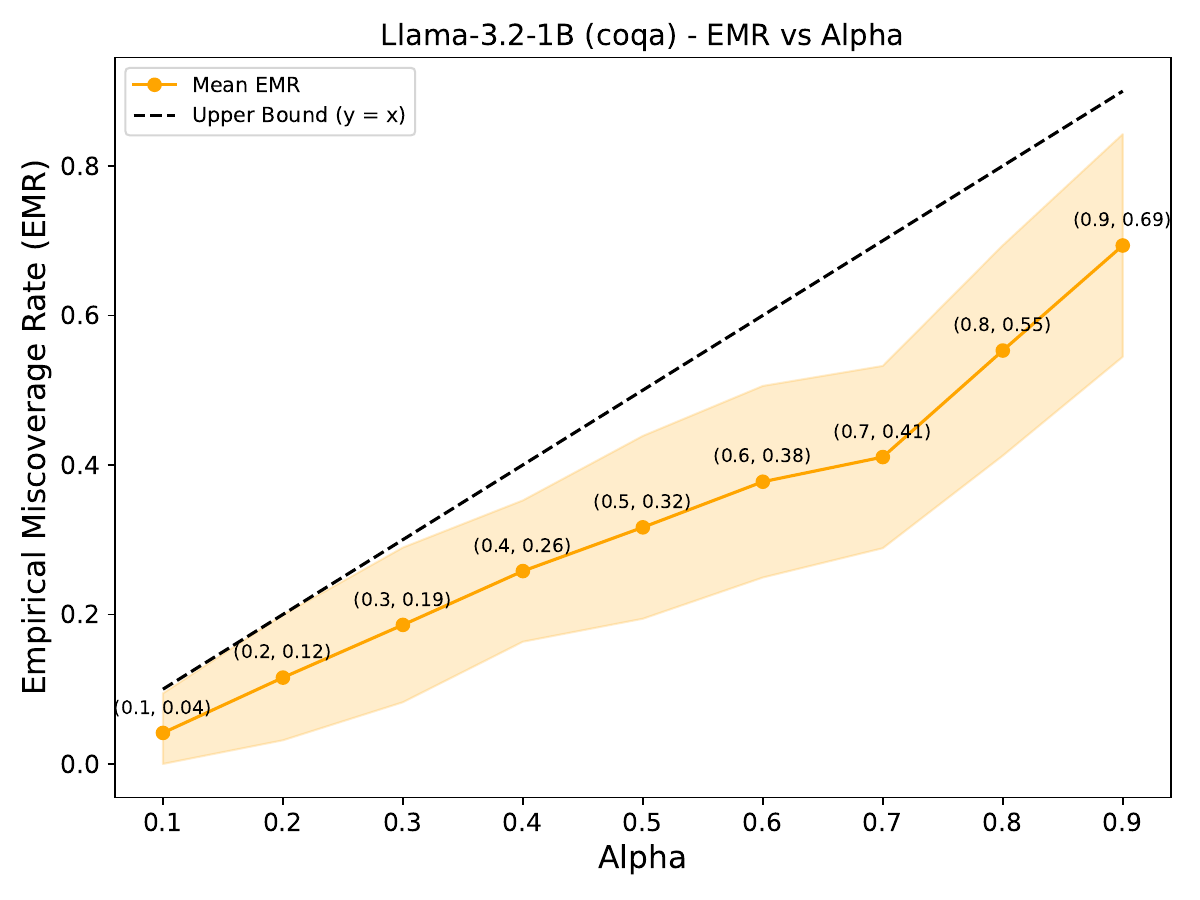}%
    }\hfill
    \subfigure[Qwen2.5-7B-Instruct]{%
        \includegraphics[width=.32\textwidth]{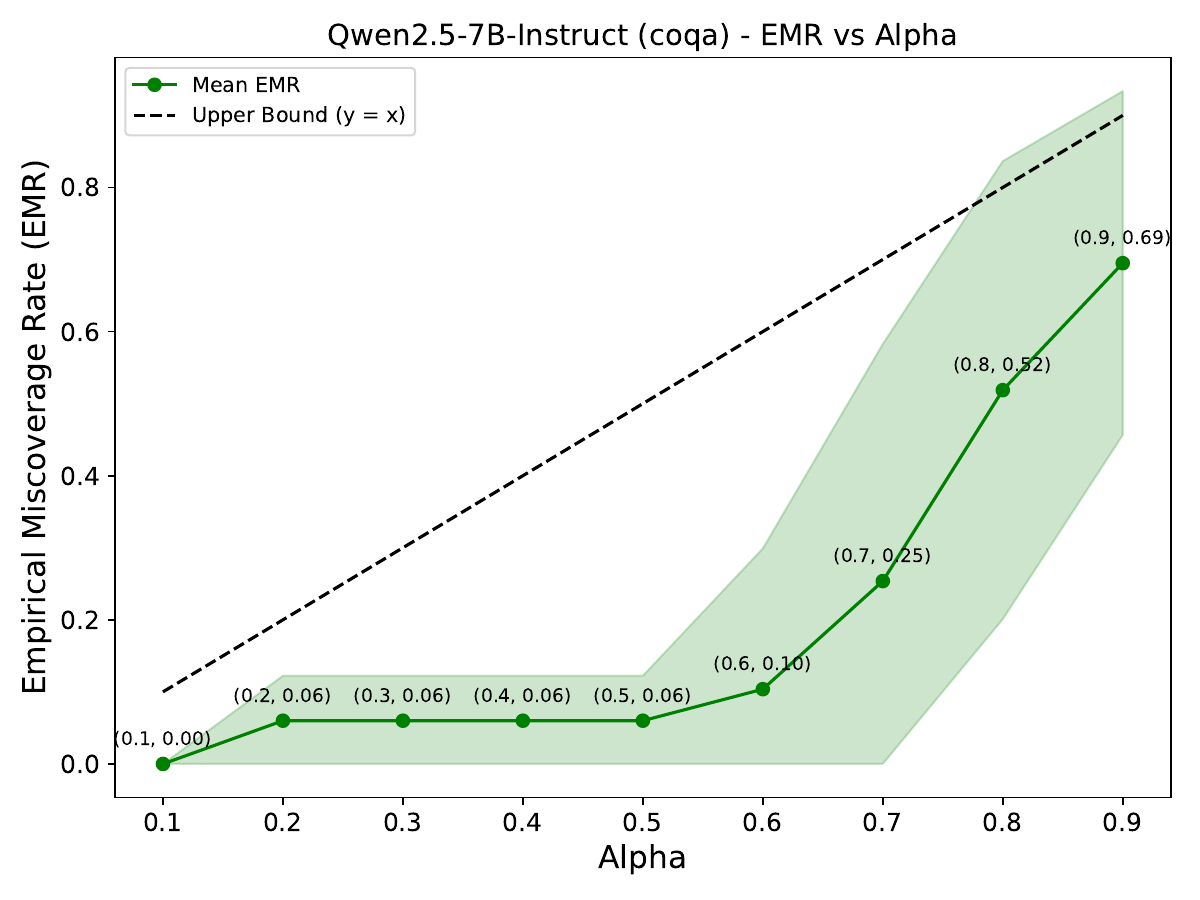}%
    }\\[-0.5em]
    \subfigure[Llama-3.1-8B-Instruct]{%
        \includegraphics[width=.32\textwidth]{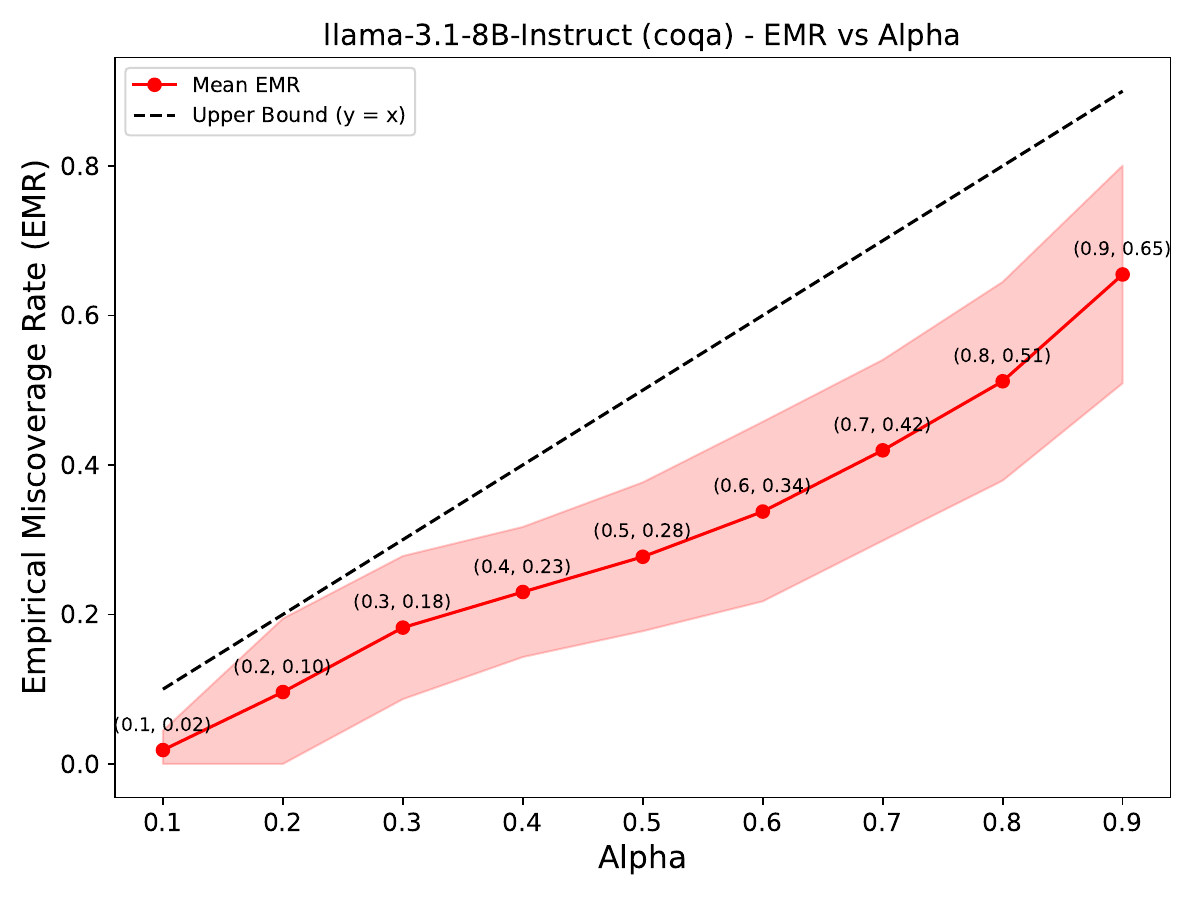}%
    }\hfill
    \subfigure[vicuna-7b-v1.5]{%
        \includegraphics[width=.32\textwidth]{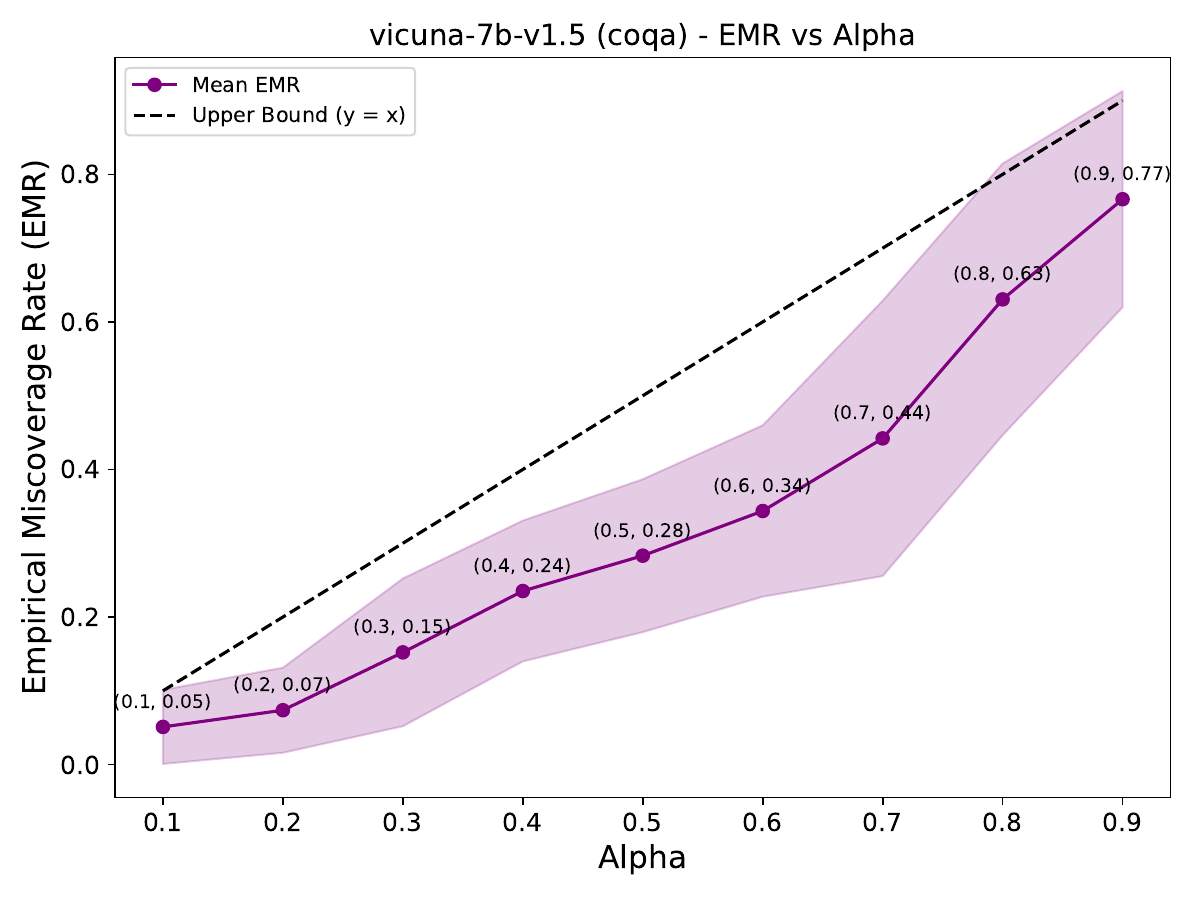}%
    }\hfill
    \subfigure[vicuna-13b-v1.5]{%
        \includegraphics[width=.32\textwidth]{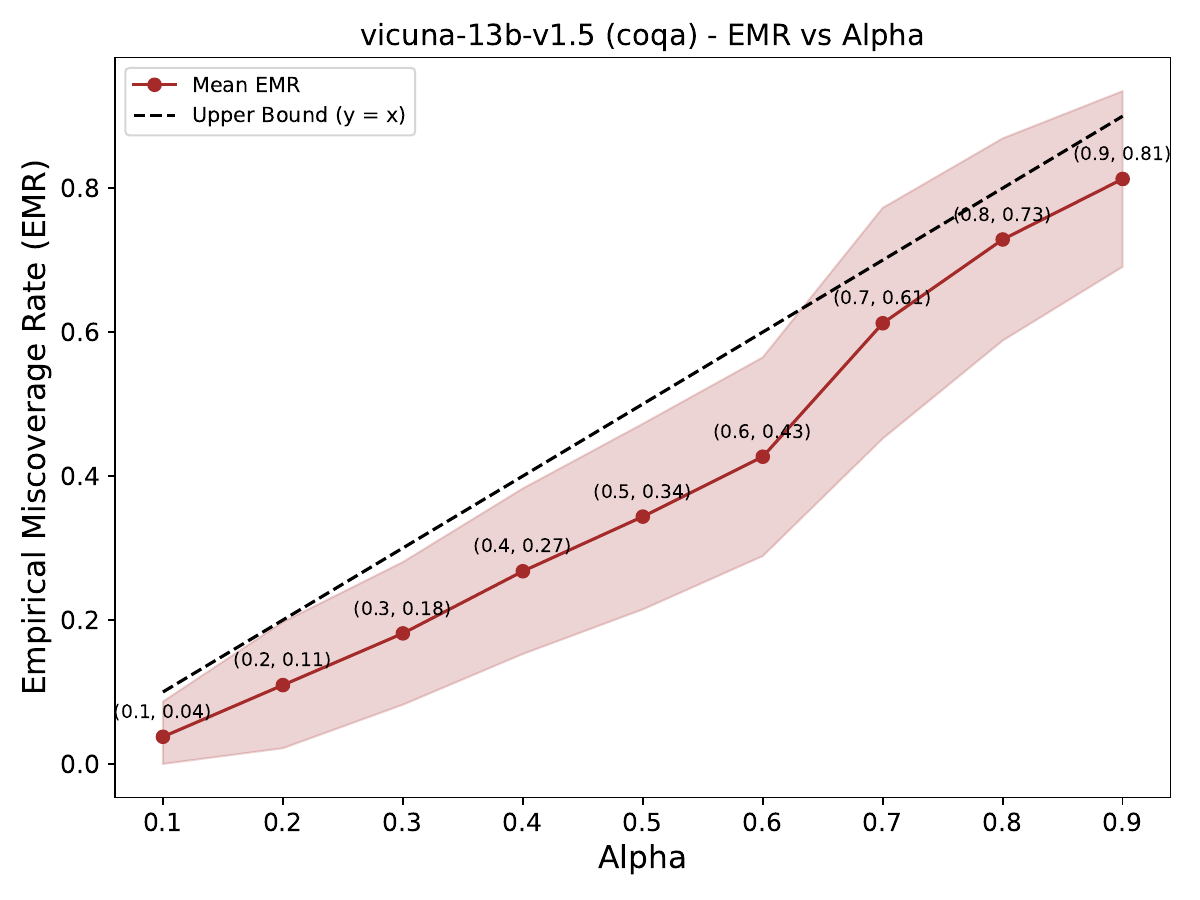}%
    }
    \caption{EMR vs. Alpha for six models on the CoQA dataset using \textbf{ConU}}
\end{figure*}
\subsection{Ablation Study}
To comprehensively evaluate the impact of different split ratios on the coverage performance of the prediction set, we conduct ablation experiments on two open-domain question answering datasets: TriviaQA and CoQA. In this experiment, we vary the calibration--test split ratios across three configurations: 0.3, 0.5, and 0.7, corresponding to different proportions of calibration data. All evaluations are performed under a fixed risk level of $\alpha = 0.1$, and the results are averaged over 100 random seeds to mitigate the influence of sampling variability. The resulting empirical coverage (defined as $1 - \text{EMR}$) is visualized using line charts in Figures 1 and 3, which illustrate model-wise performance under each split ratio.

When reducing the split ratio, risk control remains intact, as the test-time average coverage rate consistently exceeds $1 - \alpha$, demonstrating the robustness of our method. Furthermore, the efficiency of the approach is highlighted by its ability to achieve reliable guarantees on a large test set using only a limited amount of calibration data. These results suggest that the method is both robust and practical under varying data partitioning schemes.

From a cross-dataset perspective, despite differences in language style, task formulation, and reasoning complexity, our method demonstrates strong transferability and consistent coverage control across both data distributions. Under all three split ratio settings, the majority of models achieve stable coverage within the narrow interval $[0.94, 1.00]$, with no significant degradation or volatility observed. Notably, Vicuna-13B maintains perfect coverage ($1.00$) across all splits on CoQA, while LLaMA-3.1-8B and Qwen2.5-7B consistently reach coverage above $0.99$, indicating that the method reliably calibrates confidence even with limited calibration data.

\begin{table*}[!t]
\centering
\setlength{\tabcolsep}{4pt}
\renewcommand{\arraystretch}{1.1}
\resizebox{\textwidth}{!}{
\begin{tabular}{llccccccccc}
\toprule
Dataset & LLMs/$\alpha$ & 0.1 & 0.2 & 0.3 & 0.4 & 0.5 & 0.6 & 0.7 & 0.8 & 0.9 \\
\midrule
\multirow{6}{*}{TriviaQA} & Llama-3.1-8B-Instruct        & 9.02 & 8.05 & 7.06 & 6.10 & 5.10 & 4.06 & 3.03 & 2.00 & 1.02 \\
& Llama-3.2-1B                 & 9.00 & 7.99 & 6.99 & 6.00 & 5.00 & 4.00 & 3.01 & 1.99 & 1.01 \\
& Qwen2.5-7B-Instruct          & 9.01 & 8.03 & 7.02 & 6.00 & 4.97 & 3.95 & 2.96 & 2.00 & 1.00 \\
& Qwen2.5-3B-Instruct          & 8.99 & 7.99 & 7.00 & 6.00 & 4.99 & 4.00 & 3.02 & 2.03 & 1.05 \\
& vicuna-7b-v1.5               & 9.01 & 8.00 & 6.99 & 6.02 & 5.02 & 4.00 & 3.00 & 2.00 & 1.01 \\
& vicuna-13b-v1.5              & 9.00 & 7.98 & 6.97 & 5.98 & 4.97 & 3.99 & 3.00 & 1.99 & 1.02 \\
\midrule
\multirow{6}{*}{CoQA} & Llama-3.1-8B-Instruct        & 9.02 & 8.02 & 7.01 & 6.02 & 5.03 & 4.03 & 3.03 & 2.01 & 1.00 \\
& Llama-3.2-1B                 & 9.00 & 7.99 & 7.01 & 6.01 & 5.01 & 4.03 & 3.03 & 2.02 & 1.02 \\
& Qwen2.5-7B-Instruct          & 8.95 & 8.01 & 7.03 & 6.01 & 5.02 & 4.03 & 3.04 & 2.05 & 1.01 \\
& Qwen2.5-3B-Instruct          & 8.95 & 7.96 & 6.99 & 6.02 & 5.03 & 4.02 & 3.00 & 2.03 & 1.01 \\
& vicuna-7b-v1.5               & 9.00 & 8.00 & 7.00 & 6.01 & 5.01 & 4.01 & 3.00 & 2.00 & 1.00 \\
& vicuna-13b-v1.5              & 8.98 & 7.98 & 6.98 & 5.98 & 4.98 & 3.98 & 2.98 & 2.00 & 1.02 \\
\bottomrule
\end{tabular}}
\caption{Result of the prediction set size at various risk levels}
\label{tab:predset_size_vs_alpha}
\end{table*}

\begin{figure*}[!t]
\centering
\begin{minipage}{0.48\textwidth}
    \centering
    \includegraphics[width=\linewidth]{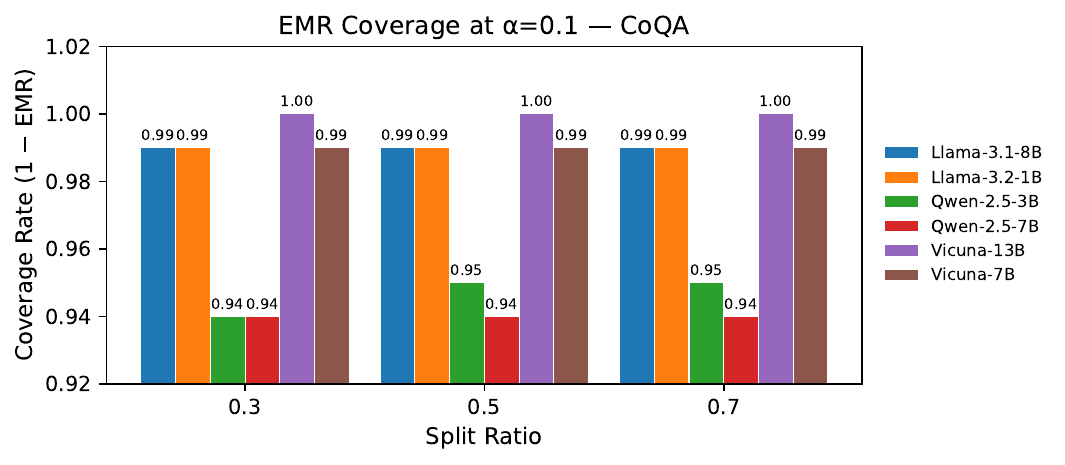}
    \vspace{-12pt}
    \caption{CoQA}
\end{minipage}
\hspace{0.02\textwidth}
\begin{minipage}{0.48\textwidth}
    \centering
    \includegraphics[width=\linewidth]{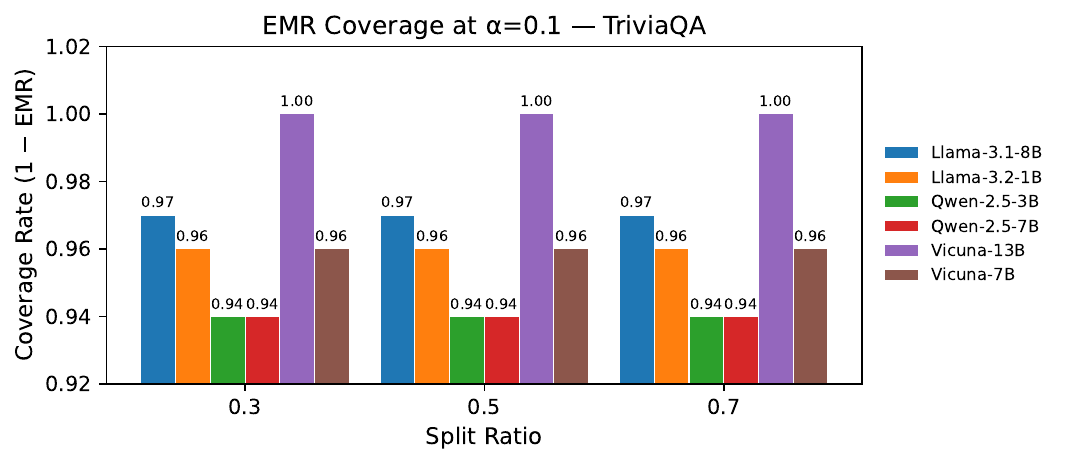}
    \vspace{-12pt}
    \caption{TriviaQA}
\end{minipage}
\end{figure*}

A finer-grained analysis of the intra-figure bar distributions further reveals that larger models exhibit minimal sensitivity to split ratio variation, whereas mid-sized and smaller models such as Qwen2.5-3B and Vicuna-7B show mild fluctuations (minimum coverage around $0.94$), yet without clear degradation trends or anomalous behavior. Additionally, the height of the bars within each group remains closely aligned, with no conspicuous stratification, reinforcing the conclusion that the conformal prediction framework—when combined with uncertainty-based ranking—exhibits strong robustness to changes in calibration--test data partitioning.

\subsection{Comparison with ConU}

By comparing our proposed method (TECP) with the ConU baseline, the results on the TriviaQA dataset are shown in Figures 1 and 2, and the results on the CoQA dataset are shown in Figures 3 and 4. We find that TECP consistently produces empirical miscoverage rate (EMR) curves with lower variance across different random seeds and calibration-test splits, indicating stronger stability. Furthermore, while ConU exhibits wider confidence intervals and higher fluctuations across multiple models, TECP demonstrates closer alignment with the theoretical upper bound $y = \alpha$ and significantly narrower uncertainty bands across all models. This suggests that TECP maintains consistent performance under random partitioning and better achieves the target theoretical risk control.

Token entropy captures the internal uncertainty of the language model by modeling the probability distribution over tokens in an autoregressive generation process. In contrast, frequency-based approaches assess semantic diversity solely from sampled outputs and are thus vulnerable to issues such as hallucination. Specifically, when a language model is highly confident yet repeatedly generates an incorrect answer, frequency-based uncertainty estimates may misleadingly appear low, failing to reflect the model’s true epistemic uncertainty. This discrepancy introduces bias into the construction of nonconformity scores, undermining the reliability of the measured disagreement between input and output. As a result, when calibration and test sets are randomly split, such errors are further amplified, leading to large variance and unstable uncertainty quantification.

\section{Conclusion}
We propose and validate a method for prediction set construction that leverages uncertainty-based ranking in conjunction with conformal calibration to achieve rigorous coverage control. Comprehensive evaluations on the TriviaQA and CoQA datasets—spanning multiple models and calibration–test splits—demonstrate that the approach maintains stable coverage and compact prediction set sizes across variations in model scale, task distribution, and data partitioning. As model capacity increases, the resulting prediction sets exhibit greater concentration and selectivity; meanwhile, coverage remains robust under different split ratios, with all findings averaged over repeated random trials to ensure statistical stability. Overall, this white-box framework achieves provable coverage and strong transferability, mitigates calibration bias in uncertainty scores, and offers a practical solution for reliable prediction and risk control in large language model applications.

\section*{Limitations}
Our approach reflects a typical limitation of white-box conformal prediction frameworks—namely, the reliance on uncertainty scores derived from model output probabilities. Although predictive uncertainty supports the creation of prediction sets with theoretical coverage guarantees, the effectiveness of calibration largely depends on the stability and accuracy of the underlying uncertainty signal, which can be affected by instruction tuning or reinforcement learning. Furthermore, the method presumes that such uncertainty scores can be reliably extracted across diverse models and tasks, an assumption that may break down under distribution shifts or non-standard decoding protocols. Future directions include developing scoring functions that are more robust to model-specific artifacts, or exploring hybrid frameworks that integrate both internal and external uncertainty cues to improve generalization.

\nocite{langley00}

\bibliography{references}
\bibliographystyle{icml2023}

\newpage
\appendix
\onecolumn


\end{document}